\documentclass[a4paper,fleqn]{cas-dc}
\usepackage[numbers,sort&compress]{natbib}
\usepackage{amssymb}
\usepackage{graphicx}
\usepackage{caption}
\usepackage{stmaryrd}
\usepackage{latexsym,amssymb}
\usepackage{amsmath,amsbsy}
\usepackage{amsopn,amstext}
\usepackage{bm}
\usepackage{float}
\usepackage{picins}
\usepackage{subfigure}
\usepackage{graphicx}
\usepackage{color}
\usepackage{longtable}
\usepackage{booktabs}
\usepackage{multirow}
\usepackage{graphicx}
\usepackage{threeparttable}
\usepackage{makecell}
\usepackage{tabularx}
\usepackage{pdfpages}

\usepackage{multirow}
\usepackage{algorithm}
\usepackage{algpseudocode}
\usepackage{hyperref}

\definecolor{hyperref-blue}{RGB}{0, 0, 0}
\newcommand{\reffig}[1]{Fig. \ref{#1}}
\newcommand{\refeq}[1]{Eq. \ref{#1}}
\newcommand{\reftable}[1]{Table \ref{#1}}

\def\tsc#1{\csdef{#1}{\textsc{\lowercase{#1}}\xspace}}
\tsc{WGM}
\tsc{QE}
\tsc{EP}
\tsc{PMS}
\tsc{BEC}
\tsc{DE}

\begin{document}
\let\WriteBookmarks\relax
\def\floatpagepagefraction{1}
\def\textpagefraction{.001}

\shorttitle{}    
\shortauthors{Junkai Mao et ~ al.} 
\title [mode = title]{MPSTAN: Metapopulation-based Spatio-Temporal Attention Network for Epidemic Forecasting}  
%
\author[1]{Junkai Mao}[type=editor ]
\ead{maojk@shu.edu.cn}
\credit{Investigation, Methodology, Modeling, Simulation, Validation, Writing - original draft}
%
%
\author[1,2,3]{Yuexing Han}[type=editor, orcid=0000-0002-1170-202X, ]
\cormark[1]
\ead{han_yx@i.shu.edu.cn}
\cortext[cor1]{Corresponding author}
\credit{Review, Editing}
%
\author[1]{Bing Wang}[type=editor, orcid=0000-0002-7078-4352, ]
\cormark[2]
\ead{bingbignwang@shu.edu.cn}
\cortext[cor1]{Corresponding author}
\credit{Modeling, Validation, Review, Editing}
%
\address[1]{School of Computer Engineering and Science, Shanghai University, Shanghai 200444, PR China}
\address[2]{Key Laboratory of Silicate Cultural Relics Conservation (Shanghai University), Ministry of Education, PR China}
\address[3]{Zhejiang Laboratory, Hangzhou 311100, PR China}

\begin{abstract}
Accurate epidemic forecasting plays a vital role for governments in developing effective prevention measures for suppressing epidemics. Most of the present spatio-temporal models cannot provide a general framework for stable, and accurate forecasting of epidemics with diverse evolution trends. Incorporating epidemiological domain knowledge ranging from single-patch to multi-patch into neural networks is expected to improve forecasting accuracy. However, relying solely on single-patch knowledge neglects inter-patch interactions, while constructing multi-patch knowledge is challenging without population mobility data. To address the aforementioned problems, we propose a novel hybrid model called Metapopulation-based Spatio-Temporal Attention Network (MPSTAN). This model aims to improve the accuracy of epidemic forecasting by incorporating multi-patch epidemiological knowledge into a spatio-temporal model and adaptively defining inter-patch interactions. Moreover, we incorporate inter-patch epidemiological knowledge into both the model construction and loss function to help the model learn epidemic transmission dynamics. Extensive experiments conducted on two representative datasets with different epidemiological evolution trends demonstrate that our proposed model outperforms the baselines and provides more accurate and stable short- and long-term forecasting. We confirm the effectiveness of domain knowledge in the learning model and investigate the impact of different ways of integrating domain knowledge on forecasting. We observe that using domain knowledge in both model construction and loss functions leads to more efficient forecasting, and selecting appropriate domain knowledge can improve accuracy further.
\end{abstract} 

\maketitle

\section{Introduction}
In the past few years, COVID-19 has emerged as a significant threat to both human life and the global economy. Due to its highly contagious nature, millions of people have been infected, leading to enormous pressure on healthcare systems and social order~\cite{kaye2021economic}. Thus, it is imperative for governments and public health departments to devise effective epidemic prevention strategies, and accurate forecasting of the outbreak's future evolution is a critical factor in preventing disease transmission, mitigating its impact on public health and the economy, and enhancing the quality and efficacy of medical services~\cite{zeroual2020deep}.
\par Traditional epidemic forecasting models use compartmental models constructed from differential equations to simulate the potential transmission dynamics of epidemic at the patch level, such as the SIR model~\cite{kermack:1972}, SEIR model~\cite{efimov2021interval} and their variants~\cite{liao2020tw,lopez2021modified}. Taking the SIR model as an example, it is used to estimate the fluctuations in the number of susceptible, infected, and recovered individuals within a single patch to understand the dynamics of epidemic in a particular patch. Many traditional time-series methods can directly forecast the temporal dependency of epidemic outbreaks, such as ARIMA~\cite{alabdulrazzaq2021accuracy} and SVR~\cite{parbat2020python}. In recent years, deep learning has been widely used in the field of time series forecasting, and several excellent models have been proposed, including LSTM~\cite{hochreiter1997long}, GRU~\cite{chung2014empirical}, Transformer~\cite{vaswani2017attention}, and Neural ODE~\cite{chen2018neural}. These models are designed to effectively handle the unique properties of time series data, such as temporal correlation, periodicity, etc. 
\par However, the above methods only consider the temporal dependence of the data and ignore the spatial dependence, which may lead to insufficiently accurate forecasting results. The reason is that the epidemic evolution of a patch is not only influenced by its own factors, such as the scale of infection and medical resources, but also by external factors, such as the mobility of people from other patches~\cite{hazarie2021interplay}. Therefore, it is crucial to consider spatial dependence to improve the accuracy of epidemiological trend analysis and forecasting. The development of graph-based algorithms provides researchers with a powerful tool for taking epidemic forecasting as a spatio-temporal forecasting problem~\cite{kipf2017semisupervised,veličković2018graph}. Various methods~\cite{kapoor2020examining,deng2020cola,zhang2021multi} have been proposed for epidemic spatio-temporal forecasting. In essence, these methods construct a graph to predict multi-patch epidemics. Each patch is represented as a node, and each patch's historical data, such as the infected cases, recovered cases, hospitalizations, and ICU admissions, are used as node features. By modeling the temporal and spatial dependencies in epidemic data, these methods can capture potential spatio-temporal correlations to predict future trends in the epidemic spreading. With the benefit of spatio-temporal forecasting works in the traffic flow field, most of the spatio-temporal models can also be directly applied to epidemic forecasting, such as~\cite{yu2018spatio,li2018diffusion,wu2019graph}.
\par
\begin{figure*}[t]
	\centering  
	\subfigure[US active cases]{
		\includegraphics[width=0.48\linewidth]{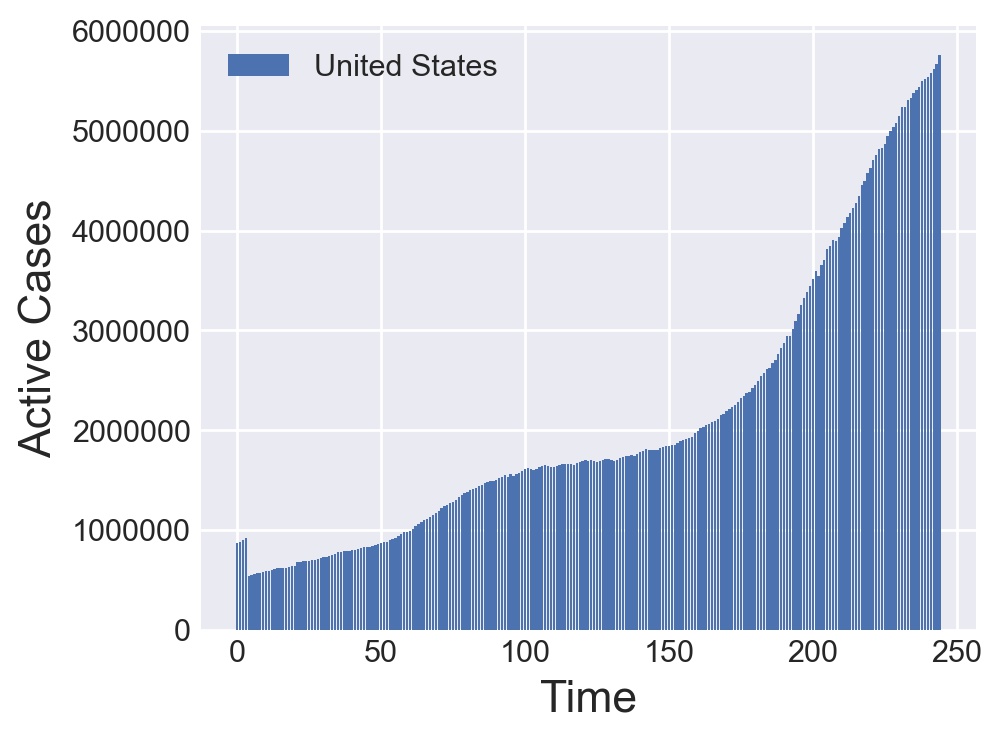}}
	\subfigure[Japan active cases]{
		\includegraphics[width=0.48\linewidth]{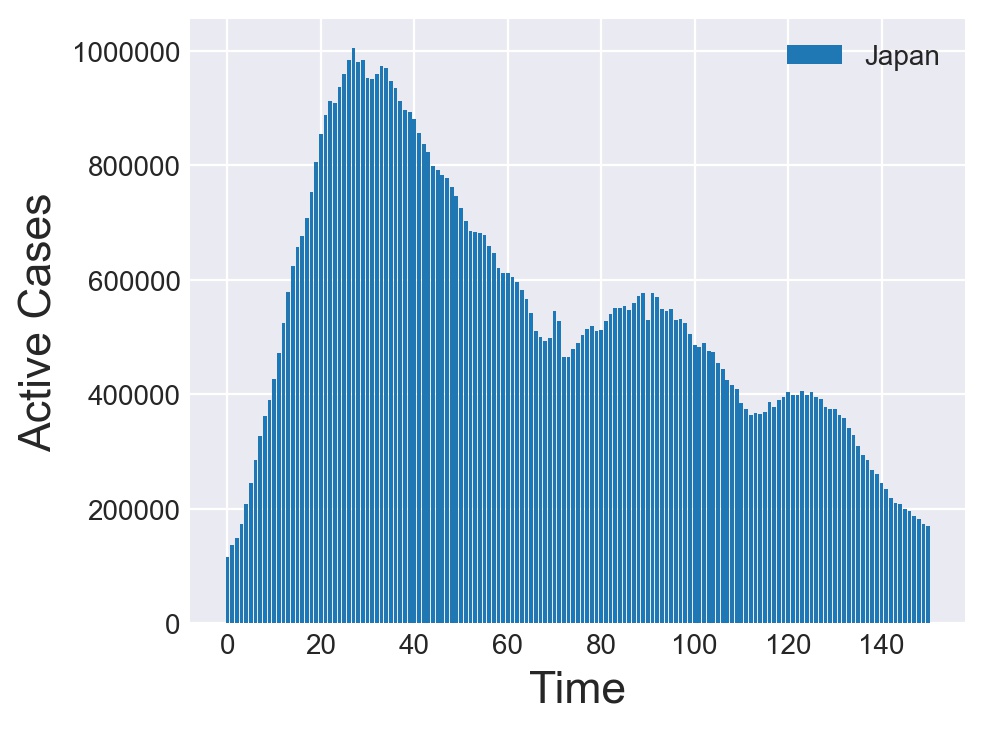}}
	\caption{Illustration of active cases on the US and Japan datasets.}
    \label{fig:all}
\end{figure*}

 Nevertheless, epidemiological evolution trends can vary considerably depending on the timing, region, and preventive measure of the epidemic outbreak. We show the number of active cases in the United States and Japan at different recording times in \reffig{fig:all}, respectively. As shown in \reffig{fig:all}, these two datasets show completely different epidemiological evolution trends. \reffig{fig:all}(a) indicates that the outbreak is ongoing, and \reffig{fig:all}(b) indicates that the outbreak is under control, where the different trends reflect the vastly different transmission dynamics of the epidemic. Traditional spatio-temporal models only find a nonlinear mapping between input and output data, without the underlying physical information, which also makes it difficult to provide stable and accurate forecasting in the face of complex trends~\cite{adiga2022ai}. In response to this issue, ~\cite{kamalov2022deep} points out that it is not reasonable to simply apply deep learning to epidemic forecasting. Furthermore, theory-guided data science demonstrates that incorporating domain knowledge into data-driven models helps improve algorithm performance~\cite{karpatne2017theory}. Therefore, researchers have attempted to use epidemiological domain knowledge to help models better learn the underlying dynamics of epidemics. Some works, such as~\cite{la2020epidemiological,gao2021stan,wang2022causalgnn}, incorporate single-patch epidemic models such as SIR and SIRD into spatio-temporal models, providing meaningful epidemiological context for neural networks and improving the performance of epidemic forecasting. However, they neglect inter-patch epidemic transmission, so some researchers~\cite{Cao2022MepoGNNME} use population mobility data to construct a metapopulation epidemic transmission model and train the learning model using this domain knowledge.


\par Although existing methods have achieved success in this field, we find the following issues:
\begin{enumerate}[(1)]
\item Most of the existing methods fail to make full use of the more reasonable epidemiological domain knowledge to help model training. They utilize domain knowledge that either ignores inter-patch interactions~\cite{la2020epidemiological,gao2021stan} or requires additional population mobility data to construct inter-patch interactions~\cite{Cao2022MepoGNNME}. The latter approach relies heavily on population mobility data, but collecting population mobility data between patches is inherently challenging and inaccurate, which can also bias the model.
\item Most of the existing domain knowledge-based models do not analyze the effectiveness of domain knowledge on model training in detail. Most methods only apply epidemiological domain knowledge to the loss function~\cite{gao2021stan,Cao2022MepoGNNME}, and some works apply epidemiological knowledge to the model construction at the same time~\cite{wang2022causalgnn}. However, these methods do not analyze in detail the effectiveness of domain knowledge on model construction and loss function separately for epidemic forecasting.
\end{enumerate}  

\par To address the above issues, we propose a novel approach named Metapopulation-based Spatio-Temporal Attention Network (MPSTAN). MPSTAN employs the MP-SIR model that considers inter-patch mobility to help spatio-temporal model training. Specifically, the MP-SIR physical model utilizes the neural network to learn physical model parameters both intra- and inter-patch, thus enabling adaptive construction of interactions between patches. Furthermore, we believe that different parameters are influenced by distinct types of information. The intra-patch parameters primarily represent the scale of the epidemic within a given patch, which reflects the temporal variations in population size for each state. The inter-patch parameters, on the other hand, capture the population mobility between patches and are also influenced by spatial information. Therefore, we design multiple parameter generators to solve the intra- and inter-patch parameters using the data containing different information as input, respectively. 
In addition, we apply the physical model to the model construction and loss function of the MPSTAN model, and thoroughly analyze the effectiveness of different ways of combining the physical model with the learning model for epidemic forecasting. Furthermore, single physical model do not accurately represent the potential epidemiological dynamics in various real-world environments. To make more accurate forecasting, selecting an appropriate epidemiological physical model tailored to the specific circumstances is necessary. In summary, the main contributions of this paper are as follows:
\begin{enumerate}[(1)]
  \item We propose a novel spatio-temporal epidemic forecasting model that employs an adaptive approach to construct the metapopulation epidemic transmission and integrates domain knowledge to aid neural network training. This spatio-temporal model does not rely on population mobility data and is capable of accurately predicting epidemic transmission.
  \item We design multiple parameter generators to learn the physical model parameters for intra- and inter-patch separately. Due to the fact that different parameters represent different information, we utilize embedding representations containing diverse information to feed into each parameter generator separately, in order to learn the corresponding physical model parameters.
  \item We reveal the significance of epidemiological domain knowledge in spatio-temporal epidemic forecasting by comparing its different incorporation methods into neural networks. Also, we emphasize the crucial importance of selecting an appropriate domain knowledge to simulate the potential epidemic transmission within the actual circumstances.
  \item We conduct extensive experiments to validate the performance of MPSTAN on two datasets with different epidemiological evolution trends. The results show that MPSTAN makes accurate short- and long-term forecasting and has the generalization ability for different epidemic evolutions.
\end{enumerate}  
\par The remainder of this paper is structured as follows: In Section \ref{sec:related work}, we introduce the related work. Section \ref{sec:Methodologies} describes the detailed design of our proposed model. Section \ref{sec:Experiments} demonstrates the experimental results and provides an analysis of the findings. Finally, a summary of the entire work is presented in Section \ref{sec:Conclusion}.
\section{Related work} \label{sec:related work}
\par Many methods have been proposed for epidemic forecasting, which are divided into four types of methods: traditional mathematical models, time series models, traditional spatio-temporal models, and domain knowledge-based spatio-temporal models.
\par Traditional mathematical models: Early researchers used epidemic transmission models or traditional time-series models to predict future epidemic trends.~\cite{moein2021inefficiency} uses SIR model to predict epidemics and points out that simple SIR model is not consistent with epidemic characteristics.~\cite{cooper2020sir,lopez2021modified}propose a series of variant models based on the SIR model to better adapt to complex and variable epidemic transmission. In addition, traditional time-series models can be directly used for epidemic forecasting due to the time-series nature of the data.~\cite{benvenuto2020application} predicts the prevalence and incidence of epidemics by ARIMA.~\cite{parbat2020python} utilizes SVR to fit the epidemiological data, but the presence of numerous spikes in daily data resulted in the poor fitting. The advantages of these methods lie in their simple structure and low computational cost, but this also means that it is difficult to effectively extract the potential complex nonlinear dynamics.

\par Time series models: Deep learning is widely used in time series forecasting due to its powerful nonlinear mapping capability, where RNN and its variants LSTM, and GRU are frequently applied to capture temporal dependence.~\cite{arora2020prediction,shahid2020predictions} consider epidemic forecasting as a time series forecasting problem, mainly using LSTM and its variants for epidemic forecasting, while~\cite{wang2019defsi} proposes a two-branch LSTM to aggregate different levels of epidemiological information. The attention mechanism is also commonly used for time-series forecasting, such as~\cite{li2021long} proposes a transformer-based model to predict the change in influenza cases and design a new loss function to avoid the performance degradation of the target value. In addition,~\cite{jung2021self} combines transformer with LSTM for effective short- and long-term epidemic forecasting. Time series forecasting models typically take into account only time dependence without considering spatial dependence. However, in the case of epidemic transmission, such models ignore the effect of inter-patch interactions on epidemic evolution. Thus, relying on temporal dependence alone can lead to inaccurate epidemic forecasting.

\par Traditional spatio-temporal models: Numerous studies have indicated that Graph Convolutional Network(GCN) show superior results in processing data with spatial structure~\cite{wu2020comprehensive,bui2022spatial}, and epidemic transmission can automatically be translated into graph structure due to its spatial nature~\cite{panagopoulos2021transfer,tomy2022estimating}.~\cite{kapoor2020examining} uses time series data as input to GCN for epidemic forecasting.
~\cite{deng2020cola} proposes a dynamic location-aware attention mechanism to capture the spatial relationships between patches. Furthermore,~\cite{zhang2021multi} fuses multimodal information in a spatio-temporal model to explore regional correlations in the epidemic transmission process. Due to the inherent nature of spatio-temporal features, models from other domains can also be applied to epidemic forecasting, such as~\cite{wu2019graph} proposes adaptive adjacency matrices to learn the relationships between nodes in a graph,~\cite{chen2021graph} chooses to model the temporal and spatial dimensions in parallel, since the complex mapping of serial neural network structures may cause the original spatio-temporal relationships to change, and~\cite{fang2021spatial} combines Neural ODE with GCN, proposes a tensor-based model that models the spatio-temporal dependencies simultaneously to avoid limiting the model representation capability. Nevertheless, traditional spatio-temporal models lacking physical information are difficult to fit the potentially complex dynamics~\cite{wang2022predicting}.

\par Domain knowledge-based spatio-temporal models: Several works have incorporated domain knowledge from epidemiology into neural networks.~\cite{la2020epidemiological} utilizes a spatio-temporal model to predict the infection rates and combines them with the SIR model to predict infected cases.~\cite{gao2021stan} constructs a physical-guided dynamic constraint model which uses the SIR model to constrain the propagation dynamics in neural network forecasting. This dynamic constraint is based on the infection and recovery rates, as well as the previous moment data, to recursively derive the predicted values. Moreover,~\cite{wang2022causalgnn} proposes a causal encoder-decoder structure based on the SIRD model, which applies not only to the loss function, but also iteratively for model construction. However, this domain knowledge (SIRD model) neglects the interactions between patches. Additionally,~\cite{Cao2022MepoGNNME} combines population mobility data to construct a metapopulation epidemic transmission model and incorporates the domain model into a neural network to help learn potential epidemic transmission dynamics. Although, it is worth noting that the accuracy and completeness of mobility data can significantly affect its performance.

\begin{figure*}[h]
\centering
\includegraphics[width=\linewidth]{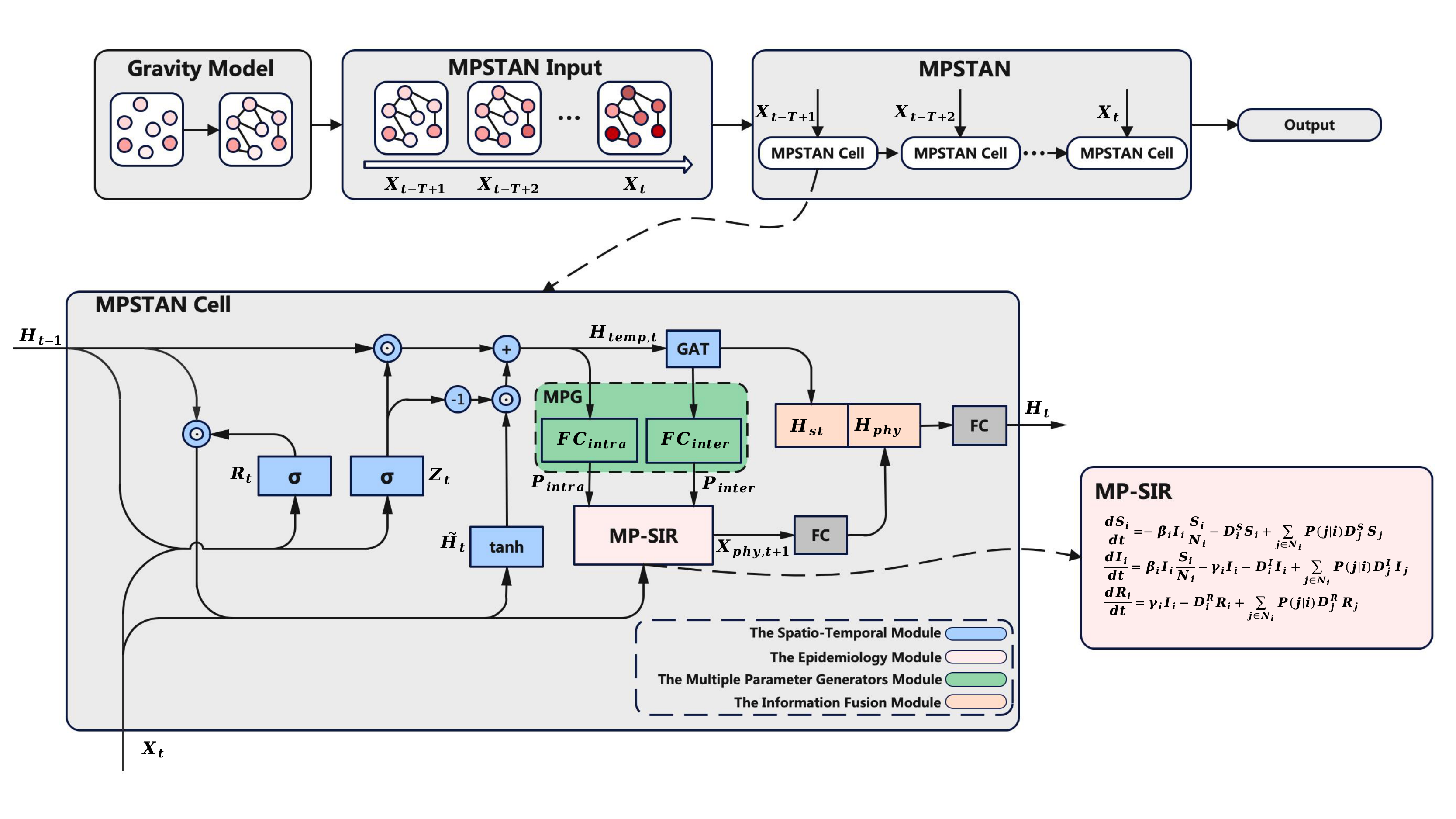}
\caption{The framework of the MPSTAN model.}
\label{fig:MPSTAN}
\end{figure*}
\section{Methodology} \label{sec:Methodologies}
In this section, we first give the problem description for epidemic forecasting. Then, we present an overview of the proposed model and details of the modules.
\subsection{Problem Description}
We use the graph $G(\mathcal{V} ,\mathcal{E} )$ to represent a spatial network, where $\mathcal{V}$ denotes the set of $N$ patches, and $\mathcal{E}$ denotes the set of edges between patches. The adjacency matrix $A \in \mathbb{R} ^{N\times N}$ represents the connections between patches. In particular, we construct the adjacency matrix by using the gravity model~\cite{truscott2012evaluating}. The edge weight between patches $i$ and $j$, $w_{ij} $ is defined as:
\begin{align}
    w_{ij}=p_{i}^{\alpha_{1}}p_{j}^{\alpha_{2}}e^{-\frac{d_{ij} }{r}}, 
\end{align}
where $p_{i}$($p_{j}$) denotes the population size of patches $i$($j$), $d_{ij}$ denotes the distance between patches $i$ and $j$. $\alpha_{1}$, $\alpha _{2}$, $r$ are the hyperparameters. It indicates that if there is high population size and closer distance between a pair of patches, there is stronger correlation epidemic propagation between the patches. We further select the maximum $E$ edge weights for all patches to make the adjacency matrix sparse, and thus reduce the computational complexity. If $w_{ij}$ belongs to the set of maximum $E$ edge weights of patch $i$, $A_{ij}=1$, otherwise $A_{ij}=0$.
\par We use $\mathcal{X} =[X_{1},X_{2},...,X_{T}]\in \mathbb{R} ^{N\times T \times C}$ to denote the spatio-temporal graph feature matrix, where $X_{t}$, $t\in \left [ 1,T \right ]$ is the graph feature matrix at time step $t$ and $C$ is the number of node features. Here, node features include the number of daily active cases, daily recovered cases, and daily susceptible cases. For epidemic forecasting, our goal is to learn a function $f(\cdot)$, which uses the adjacency matrix $A$ and the node feature matrix $X_{t-T:t} $ of historical $T$ time steps as inputs to predict the number of daily active cases $Y_{t+1:t+{T}'}$ of future ${T}' $ time steps. The problem can be formulated as follows:
\begin{align}
    [X_{t-T+1},X_{t-T+2},...,X_{t};A ]\overset{f(\cdot )}{\rightarrow}  [Y_{t+1},Y_{t+2}...,Y_{t+{T}' } ].
\end{align}
\subsection{Model Overview}
The overall framework of the MPSTAN model is shown in \reffig{fig:MPSTAN}. The model consists of a recurrent architecture and each model cell contains four modules, namely, the spatio-temporal module, the epidemiology module, the multiple parameter generators module, and the information fusion module. At first, we use the spatio-temporal module to learn the spatio-temporal information from the input data. The learned spatio-temporal information is then passed into the parameter generation module to learn the epidemiological parameters for the epidemiological model. Further, the input and the learned parameters are passed into the epidemiological module to achieve epidemic forecasting. Finally, the learned spatio-temporal information is fused with the physical forecasting information in the information fusion module, and the output containing the fused information is passed to the MPSTAN cell at the next time step.

\subsection{The Spatio-Temporal Module}
The spatio-temporal module uses the spatio-temporal feature matrix $\mathcal{X} \in \mathbb{R} ^{N\times T \times C}$ and the adjacency matrix $A \in \mathbb{R} ^{N\times N} $ to learn the spatio-temporal information of the epidemic data. This module embeds graph attention network(GAT) into gated recurrent unit(GRU), which learn spatial dependence and temporal dependence.
\paragraph{Temporal embedding.}
Initially, GRU is widely used for time series forecasting due to its ability to efficiently model time series, thus, we use GRU to learn the temporal embedding of each patch. In the GRU, $Z_{t}$, $R_{t}$ denote the update gate and reset gate at time step $t$, $\widetilde{H }_{t}$ denotes the hidden embedding at time step $t$, $H_{t-1}$ denotes the output of the MPSTAN cell at time step $t-1$, and $H_{temp,t}$ denotes the output containing the temporal dependence at time step $t$:
\begin{align}
    &Z_{t} =\sigma (W_{z}X_{t}+U_{z}H_{t-1}+b_{z} ),\\
    &R_{t} =\sigma (W_{r}X_{t}+U_{r}H_{t-1}+b_{r} ),\\
  &\widetilde{H }_{t}=\tanh (W_{h}X_{t}+U_{h}(R_{t}\odot H_{t-1}  )+b_{h} ),\\
    &H_{temp,t}=Z_{t}\odot H_{t-1}+(1-Z_{t}) \odot \widetilde{H_{t} },
\end{align}%
\noindent where $\odot $ denotes the element-wise multiplication, $W_{z}$, $W_{r}$, $W_{h}$, $U_{z}$, $U_{r}$, $U_{h}$ ,$b_{z}$, $b_{r}$, $b_{h}$ denote the learnable parameters.
\paragraph{Spatial embedding.}
The epidemic evolution of each patch is not independent, but is influenced by other patches at the spatial level. This is in a similar way as GAT, which combines attention mechanism to aggregate information from neighbor patches and update the embedding for each patch. Therefore, we use a two-layer multi-head GAT to capture the spatial dependence of epidemic evolution among patches. Firstly, we take the embedding of each patch as input and use the multi-head mechanism to compute $K$ independent attention weights. The attention weight between patch $i$ and patch $j$ at the $k$-th head as $e_{ij}^{k} $ is given by,
\begin{align}
    e_{ij}^{k}=\sigma (W_{att}^{k}((W_{temp}^{k}H_{temp,t}^{i}  )\parallel  (W_{temp}^{k}H_{temp,t}^{j}  ) )),
\end{align}%
\noindent where $W_{att}^{k}$, $W_{temp}^{k}$ denote the learnable parameters of the $k$-th head, $(\cdot \parallel  \cdot  )$ denotes the vector concatenation, $\sigma$ denotes the nonlinear activation function, and $e_{ij}^{k}$ omits the subscript $t$.
\par Then, we use the softmax function to calculate the attention scores of all the edges. The attention score between patch $i$ and patch $j$ at the $k$-th head as $a_{ij}^{k} $ is expressed as:
\begin{align}
    a_{ij}^{k}=Softmax(e_{ij}^{k} ).
\end{align}%
\par Finally, the attention scores are used to aggregate the information from neighboring patches and update the patches embeddings $H_{st}\in \mathbb{R} ^{N\times D_{st}} $, where $D_{st}$ denotes the embedding dimension of each patch. The embedding of patch $i$ as $H_{st}^{i} $ is calculated as:
\begin{align}
    H_{st}^{i} =\frac{1}{K}\sum_{k=1}^{K}  \sum_{j\in \mathcal{N}_{i}}^{} a_{ij}^{k}W^{k}_{temp}H_{temp,t}^{j},
\end{align}%
\noindent where $\mathcal{N}_{i}$ denotes the set of neighbors of patch $i$. If $A_{ij}=1$, it indicates that patch $j$ belongs to the set of neighbors of patch $i$.

\subsection{The Epidemiology Module}
We observe that the results of epidemic forecasting using only spatio-temporal models are not accurate and stable, and it is also very challenging to predict for datasets with different epidemiological evolution trends (e.g., outbreak, outbreak under control)~\cite{adiga2022ai}. Therefore, some works choose to use epidemiological domain knowledge to help model training, such as~\cite{gao2021stan,wang2022causalgnn}. These works mainly use compartmental models as domain knowledge, such as the SIR model. The SIR model is the most typical model in epidemic transmission, where S denotes the susceptible individuals, I denotes the infected individuals, and R denotes the recovered individuals. The model uses three differential equations to represent the number of changes in the three state populations in patch $i$:
\begin{align}
    &\frac{dS_{i}}{dt}=-\beta_{i}I_{i} \frac{S_{i} }{N_{i} },\\
    &\frac{dI_{i} }{dt}=\beta_{i}I_{i}\frac{S_{i} }{N_{i} }-\gamma_{i}I_{i}, \\
    &\frac{dR_{i} }{dt}=\gamma_{i} I_{i},
\end{align}%
\noindent where $\beta_{i} $ and $\gamma_{i} $ denote the infection and recovery rate of epidemic transmission in patch $i\in \left [ 1,\dots,N \right]$. However, the SIR model is limited to simulate epidemic transmission within a single patch, and neglect the inter-patch interactions. Therefore,~\cite{Cao2022MepoGNNME} uses population mobility data to construct a metapopulation epidemic model, and iteratively calculates the daily confirmed cases using neural networks. In addition, other mobility change data (e.g., GPS trajectory data) can also be used to construct the metapopulation epidemic model. However, accurate collection of mobility data is challenging, and other data may not fully reflect actual population mobility patterns.
\par To overcome the limitation of data availability, we develop an adaptive approach to define inter-patch interactions and construct a metapopulation epidemic model, named the metapopulation-based SIR (MP-SIR) model, which does not rely on mobility data. The MP-SIR model is based on the original SIR model with inter-patch mobility parameters to represent the mobility of populations at each state between patches:

\begin{align}
    &\frac{dS_{i} }{dt}=-\beta_{i}I_{i}\frac{S_{i} }{N_{i} }-D_{i}^{S} S_{i} +\sum_{j\in \mathcal{N} _{i}  }^{}P(j\mid i) D_{j}^{S} S_{j}, 
    \label{eq:13}
    \\
    &\frac{dI_{i} }{dt}=\beta_{i}I_{i}\frac{S_{i} }{N_{i} }-\gamma_{i} I_{i}-D_{i}^{I} I_{i} +\sum_{j\in \mathcal{N} _{i}  }^{}P(j\mid i)D_{j}^{I} I_{j}, \\
    &\frac{dR_{i} }{dt}=\gamma_{i} I_{i}-D_{i}^{R} R_{i} + \sum_{j\in \mathcal{N} _{i}  }^{}P(j\mid i)D_{j}^{R} R_{j},
\end{align}%
\noindent where $P(j\mid i)$ denotes the mobility probability of patch $j$ to patch $i$, and $D_{i}^{S}$, $D_{i}^{I}$, $D_{i}^{R}$ denote the mobility rates of susceptible, infected, and recovered individuals in patch $i$.
\par Taking \refeq{eq:13} as an example, the change in the number of infected individuals within patch $i$ is affected by four aspects: (\romannumeral1)susceptible individuals $S_{i}$ become infected individuals with probability $\beta_{i}$ after contact with infected individuals $I_{i}$; (\romannumeral2)infected individuals $I_{i}$ recover with probability $\gamma_{i}$; (\romannumeral3)infected individuals $I_{i}$ within patch $i$ move to other patches with the mobility rate $D_{i}^{I}$; (\romannumeral4)infected individuals $I_{j}$ from patch $j$ move toward patch $i$ with the mobility rate $D_{j}^{I}$. We simply assume that the probability of a patch migrating to other neighboring patches is equal. Formally, the mobility probability of patch $j$ to patch $i$ $P(j\mid i)$ is computed as follows:
\begin{align}
    P(j\mid i)=\frac{1}{\left |   \mathcal{N}_{j} \right | }.
\end{align}%
\par We use neural networks to generate intra- and inter-patch MP-SIR model parameters $P_{intra}=[\beta,\gamma]\in \mathbb{R} ^{N\times 2}$, $P_{inter}=[D^{S},D^{I},D^{R}]\in \mathbb{R} ^{N\times 3}$, and will describe them in detail in Section \ref{sec:MPG}. Finally, the epidemic data and the generated MP-SIR model parameters are used as inputs to the MP-SIR model for domain knowledge-based epidemic forecasting:
\begin{align}
    &\Delta X_{phy,t}=MP\mbox{-}SIR(X_{t},P_{intra},P_{inter}),\\
    &X_{phy,t+1}=X_{t}+\Delta X_{phy,t},
\end{align}%
\noindent where $\Delta X_{phy,t}\in \mathbb{R} ^{N\times 3}$ denotes the change in the number of individuals in each state at time step $t$ and $X_{phy,t+1}=\left[X_{phy,t+1}^S,X_{phy,t+1}^I,X_{phy,t+1}^R \right ]\in \mathbb{R} ^{N\times 3}$ denotes the epidemic forecasting at time step $t+1$.
\subsection{The Multiple Parameter Generators Module}\label{sec:MPG}
We use embeddings containing different information to learn intra- and inter-patch physical model parameters $P_{intra}\in \mathbb{R} ^{N\times 2}$, $P_{inter}\in \mathbb{R} ^{N\times 3}$, separately, instead of directly using embeddings containing spatio-temporal information. The intra-patch physical model parameters $\beta$, $\gamma$ indicate the epidemic evolution within a single patch and are mainly affected by the temporal dependence, while the inter-patch physical model parameters $D^{S}$, $D^{I}$, $D^{R}$ indicate the inter-patch population mobility and are mainly affected by the spatio-temporal dependence. Therefore, we generate these two types of physical model parameters by passing embeddings containing only temporal dependence and spatio-temporal dependence to the two fully connected layers, respectively:
\begin{align}
    P_{intra}&=FC_{intra}(H_{temp,t} ),\\
    P_{inter}&=FC_{inter}(H_{st} ).
\end{align}%
\subsection{The Information Fusion Module}
In this module, the information between neural network forecasting $H_{st}\in \mathbb{R} ^{N\times D_{st}} $ and physical model forecasting $X_{phy,t+1}\in \mathbb{R} ^{N\times 3}$ is fused. First, we map $X_{phy,t+1}$ to $H_{phy}\in \mathbb{R} ^{N\times D_{st}}$ using a fully connected layer which aims to keep the physical forecasting the same dimensions as the neural network forecasting,
\begin{align}
    H_{phy}=FC(X_{phy,t+1} ).
\end{align}%
\par Next, the neural network forecasting is concatenated with the physical forecasting. Finally, a fully connected layer is used to generate the final output $H_{t}\in \mathbb{R} ^{N\times D_{gru}}$ of the MPSTAN cell at time step $t$, where $D_{gru}$ denotes the dimension of GRU:
\begin{align}
    H_{t}=FC(H_{st}\parallel  H_{phy} ).
\end{align}%
\subsection{Output Layer}
The output of the MPSTAN model is divided into two parts: neural network forecasting and physical model forecasting.
\paragraph{Neural network forecasting.}
We use the final output $H_{T}\in \mathbb{R} ^{N\times D_{gru}}$ of MPSTAN as the input of a fully connected layer to predict the number of infected individuals $Y^{st}\in \mathbb{R} ^{N\times {T}'}$ in all patches for the next ${T}'$ time steps:
\begin{align}
    Y^{st}=FC_{pred}(H_{T}).
\end{align}%
\paragraph{Physical model forecasting.}
The input data from the last day and the final trained model parameters are used as inputs for the MP-SIR model to recursively predict the number of infected individuals $Y^{phy}\in \mathbb{R} ^{N\times {T}'}$ in all patches for the next ${T}'$ time steps:
\begin{align}
    &\Delta X_{phy,T}=MP\mbox{-}SIR(X_{T},P_{intra,T},P_{inter,T}),\\
    &X_{phy,T+1}=X_{T}+\Delta X_{phy,T},\\
      &\dots\nonumber \\
    &Y^{phy}=[X_{phy,T+1}^{I} ,X_{phy,T+2}^{I},\dots ,X_{phy,T+{T}'}^{I}].
\end{align}%
\subsection{Optimization}
We utilize epidemiological domain knowledge for model construction and loss functions to more effectively help MPSTAN models learn the epidemiological evolution trends. We compare the predicted values $Y^{st}$, $Y^{phy}$ of neural networks and physical models with the ground truth $\widehat{Y}$ and then optimize a MAE loss via gradient descent:
\begin{align}
    \mathcal{L} (\Theta )=\frac{1}{N\times {T}' }  \sum_{i=1}^{N}\sum_{\tau =1}^{{T}'}(\left |Y^{st}_{i,\tau }-\widehat{Y}_{i,\tau }\right |+ \left |Y^{phy}_{i,\tau }-\widehat{Y}_{i,\tau } \right | ).
\end{align}%

\begin{table*}
    \centering
    \begin{tabular}{llllllll}
        \toprule
        Dataset& Data level & Data size & Time range & Min &Max   &  Mean & Std \\
        \midrule
        US & State-level & $52\times245$ & 2020.5.1-2020.12.31& 0 & 838855& 40438& 75691  \\
        Japan &Prefecture-level & $47\times151$ & 2022.1.15-2022.6.14 & 104 & 198011 & 11458 & 21188 \\
        \bottomrule
    \end{tabular}
    \caption{Statistical information of the datasets.}
    \label{table:data}
\end{table*}

\section{Experiments} \label{sec:Experiments}
\subsection{Datasets}
Our experiments are conducted on two real-world datasets: the US dataset and the Japan dataset. As shown in \reftable{table:data}, the US dataset is state-level data collected from the Johns Hopkins University Coronavirus Resource Center~\cite{dong2020interactive}, which records the number of daily active cases, daily recovered cases, daily susceptible cases and total population for 52 states from May 1, 2020 to December 31, 2020 (245 days). The Japan dataset is prefecture-level data collected from the Japan LIVE Dashboard~\cite{su2021japan}, which records the number of daily active cases, daily recovered cases, daily susceptible cases and total population for 47 prefectures from January 15, 2022 to June 14, 2022 (151 days).
\subsection{Experimental Details}
\paragraph{Baselines.}
We compare our model with the following four kinds of baselines: (\romannumeral1)traditional mathematical models: SIR, ARIMA; (\romannumeral2)time series models: GRU; (\romannumeral3)traditional spatio-temporal models: GraphWaveNet, STGODE, CovidGNN, ColaGNN; (\romannumeral4)domain knowledge-based spatio-temporal model: STAN.
\begin{enumerate}[(1)]
  \item \textbf{SIR}~\cite{kermack:1972}: The SIR model uses three differential equations to calculate the change in the number of susceptible, infected and recovered cases in a single patch.
  \item \textbf{ARIMA}~\cite{benvenuto2020application}: The Auto-Regressive Integrated Moving Average model is widely used for time series forecasting. We use ARIMA to predict daily active cases for each patch.
  \item \textbf{GRU}~\cite{chung2014empirical}: The Gated Recurrent Unit is a variant of RNN that uses fewer parameters to implement the gating mechanism compared to LSTM.  We use a GRU for each patch separately to predict daily active cases.
  \item \textbf{GraphWaveNet}~\cite{wu2019graph}: GraphWaveNet combines adaptive adjacency matrix, diffusion convolution, and gated TCN to capture spatio-temporal dependencies.
  \item \textbf{STGODE}~\cite{fang2021spatial}: STGODE proposes a spatio-temporal tensor model by combining Neural ODE with GCN to achieve unified modeling of spatio-temporal dependencies.
  \item \textbf{CovidGNN}~\cite{kapoor2020examining}: CovidGNN uses the time series of each patch as node features and predicts epidemics using GCN with skip connections.
  \item \textbf{ColaGNN}~\cite{deng2020cola}: ColaGNN designs dynamic adjacency matrix using attention mechanism and adopts a multi-scale dilated convolutional layer for long- and short-term epidemic forecasting.
  \item \textbf{STAN}~\cite{gao2021stan}: STAN applies epidemiological domain knowledge to the loss function, which specifically constructs a dynamics constraint loss by combining the SIR model.
\end{enumerate}
\paragraph{Settings.}
We split the two datasets into training sets, validation sets and test sets at the ratio of 60\%-20\%-20\% and normalize all the data to the range (0, 1). To verify the effectiveness of the model in short- and long-time forecasting, we set the input time length as 5, the forecasting time length as 5 and 10 for short-term forecasting, and the forecasting time length as 15 and 20 for long-term forecasting. In the model, the dimensions of GRU and GAT are set to 64 and 32 respectively. Besides, the number of heads K in GAT is set to 2. We set epoch numbers as 50 and use Adam optimizer with the learning rate 1e-3.

\begin{table*}
    \centering
    \resizebox{\textwidth}{!}{
    \begin{tabular}{lllllllllll}
      \toprule
        &\multicolumn{10}{l}{The US dataset}\\
        \cmidrule(l){2-11}
        &\multicolumn{5}{l}{T=5}&\multicolumn{5}{l}{T=10}\\
        \cmidrule(l){2-6}\cmidrule(l){7-11}
         Model & MAE & RMSE &MAPE & PCC & CCC  & MAE & RMSE &MAPE & PCC & CCC  \\
        \midrule 
        SIR  &\underline{5660}  & \underline{15656} &25.81\%  & 99.16\%&  \underline{99.14\% } & 10608 &  33766& 41.94\% &  96.23\% & 96.09\%  \\
        ARIMA &6475  &22095  & 14.01\% &98.33\%  &98.31\%  &11489  & 44779 &26.36\%  & 93.66\% & 93.39\% \\
        GRU & 18348 &32950  & 21.88\% & 97.88\% & 95.63\% & 26749 & 47328 &32.52\%  &95.66\%  &90.39\%   \\
        GraphWaveNet & 13875 &22559  & 17.85\% & \underline{99.46\%} & 97.82\% & \underline{9526} & \underline{15673} & 16.64\% & \underline{99.21\% }& \underline{99.09\% }\\
        STGODE & 70454 &116865  &  83.21\%& 91.95\% & 64.48\% &53693  &83823  & 63.51\% & 87.89\% &  62.19\%\\
        CovidGNN & 9453 & 21612 &\underline{9.91\%}  & 99.07\% &  98.17\%& 16052 &37586  & \underline{15.03\%} & 96.87\% & 94.00\% \\
        ColaGNN & 66005 &  111622&   77.57\% &  53.54\%& 41.79\% & 51822 & 91680 & 57.61\% & 80.18\%&62.46\%  \\
        STAN  &10024  & 19214 & 17.98\% &98.70\% & 98.65\% & 13993 &25963  &  19.38\%& 97.80\% & 97.49\% \\
        MPSTAN & \textbf{3960} & \textbf{8255} & \textbf{6.38\% }& \textbf{99.80\%} & \textbf{99.75\%} &\textbf{7711}  & \textbf{14463} &  \textbf{10.73\%}& \textbf{99.55\%} & \textbf{99.20\%} \\
        Improvement &30.04\%  & 47.27\% &35.62\%  & 0.34\% &0.61\%  & 19.05\% & 7.72\% & 28.61\%& 0.34\% &  0.11\%\\
        \toprule
        &\multicolumn{5}{l}{T=15}&\multicolumn{5}{l}{T=20}\\
        \cmidrule(l){2-6}\cmidrule(l){7-11}
          Model& MAE & RMSE &MAPE & PCC & CCC  & MAE & RMSE &MAPE & PCC & CCC  \\
        \midrule 
        SIR  &\underline{16573}  & 60984 &57.38\%  &89.04\%  &88.26\%  &23963  &101612  &76.12\%  & 76.44\% & 73.21\% \\
        ARIMA &  17151& 74295 & 43.01\% &84.86\%  & 83.59\% & 24849 &121875  & 65.20\% & 69.78\% & 65.31\% \\
        GRU & 33968 & 59804 & 41.21\% & 92.67\% & 83.94\% & 38202 & 65762 & 45.54\% & 90.44\% &80.61\%   \\
        GraphWaveNet &47020  &76735  & 51.64\% & 90.76\% & 72.64\% & 48154 &82098  & 51.19\% & 84.43\% & 68.47\% \\
        STGODE &72622  &117611  & 107.65\% & 82.26\% & 50.75\% &72132  & 109536 & 84.84\% & 85.16\% & 42.81\%  \\
        CovidGNN & 21660 &48169 & \underline{19.85\%} & 94.68\% & 89.71\% &26985  &57085  & \underline{24.57\%} & 92.64\% & 84.95\%  \\
        ColaGNN &33419  & 55424 & 41.36\% &92.43\%  &79.63\%  &47837  & 77656 & 52.48\% & 92.49\% & 70.90\% \\
        STAN  & 16784 &\underline{33383}  &20.78\%  &\underline{96.43\%}  &  \underline{95.49\%}& \underline{18679 }& \underline{36180} & 26.81\% & \underline{96.09\%} &  \underline{94.52\%} \\
        MPSTAN &  \textbf{10148}&  \textbf{18460}& \textbf{14.68\%} & \textbf{99.25\% }&  \textbf{98.68\%}  & \textbf{12728 }& \textbf{22923} & \textbf{18.68\%} & \textbf{98.81\%} & \textbf{97.91\%}\\
        Improvement &38.77\%  & 44.70\% & 26.05\% & 2.92\% & 3.34\% & 31.86\% &36.64\%  & 23.97\% &  2.83\%& 3.59\% \\
          \bottomrule
    \end{tabular}
    }
    \caption{Performance comparison with baseline on the US dataset.}
    \label{table:us-exp}
\end{table*}

\paragraph{Evaluation Metrics.}
In this study, we chose Mean Absolute Error (MAE), Root Mean Squared Error (RMSE), Mean Absolute Percentage Error (MAPE), Pearson’s Correlation Coefficient (PCC), and Concordance Correlation Coefficient (CCC) to evaluate the performance of each model, where the lower MAE, RMSE, and MAPE, and the higher PCC and CCC, the better the forecasting performance. The above evaluation metrics are expressed as follows:
\begin{align}
    &MAE=\frac{1}{N\times {T}'} \sum_{i=1}^{N}\sum_{\tau=1}^{{T}'}  (|Y_{i,\tau}^{st} -\widehat{Y}_{i,\tau } |),\\
    &RMSE=\sqrt{\frac{1}{N \times {T}' } \sum_{i=1}^{N}\sum_{\tau=1}^{{T}'}(|Y_{i,\tau}^{st} -\widehat{Y}_{i,\tau } |)^{2} }, \\
    &MAPE=\frac{100\%}{N \times {T}'} \sum_{i=1}^{N}\sum_{\tau=1}^{{T}'}|\frac{Y_{i,\tau}^{st} -\widehat{Y}_{i,\tau }}{\widehat{Y}_{i,\tau }} | , \\
    &PCC=\frac{\sum_{i=1}^{N}\sum_{\tau=1}^{{T}'}(Y_{i,\tau}^{st}-\bar{ Y}_{i,\tau}^{st} )(\widehat{Y}_{i,\tau}
-\bar{\widehat{Y}}_{i,\tau })}{\sqrt{\sum_{i=1}^{N}\sum_{\tau=1}^{{T}'} (Y_{i,\tau}^{st}-\bar{ Y}_{i,\tau}^{st} )^{2}(\widehat{Y}_{i,\tau}
-\bar{\widehat{Y}}_{i,\tau })^{2}} },\\
    &CCC=\frac{2\rho \sigma _{x}\sigma _{y}}{\sigma _{x}^{2}+\sigma _{y}^{2}+(\mu _{x}-\mu _{y})^{2}} ,
\end{align}
where $\rho$ denotes the correlation coefficient between the two variables, $\mu _{x}$ and $\mu _{y}$ denote the mean of the two variables, and $\sigma _{x}^{2}$, $\sigma _{y}^{2}$ are the corresponding variances.

\begin{table*}
    \centering
    \resizebox{\textwidth}{!}{
    \begin{tabular}{lllllllllll}
      \toprule
        &\multicolumn{10}{l}{The Japan Dataset}\\
        \cmidrule(l){2-11}
        &\multicolumn{5}{l}{T=5}&\multicolumn{5}{l}{T=10}\\
        \cmidrule(l){2-6}\cmidrule(l){7-11}
          Model& MAE & RMSE &MAPE & PCC & CCC  & MAE & RMSE &MAPE & PCC & CCC  \\
        \midrule 
        SIR  & \textbf{896} &  \textbf{1572}& 18.89\% & \textbf{99.11\% }&  \textbf{97.91\%}&  1703& \textbf{2874} & 39.38\% &\textbf{97.73\%}  & \textbf{93.67\%}  \\
        ARIMA & 1113 & 3137 & 24.33\% & 91.74\% & 91.37\% & 2433 & 8719 & 59.59\% & 63.42\% & 57.19\%  \\
        GRU   & 2156 & 3955 &58.91\%  &94.06\%  &89.02\%  & 2702 & 5130 & 69.49\% & 92.33\% & 83.80\% \\
        GraphWaveNet  & 2048 &4490  & 39.06\% & 94.93\% &87.35\%  & 2744 & 6447 &  48.88\%& 92.64\% & 79.24\% \\
        STGODE  &  5420& 13057 &103.14\%  & 83.94\% &57.16\%  &  8208&  18396& 158.08\% &85.00\%  &50.91\%   \\
        CovidGNN  &1042  & \underline{2305} & \underline{18.06\%} & \underline{97.27\%} & \underline{95.71\%} &1887  & 3942 &39.40\%  &\underline{95.77\% } & 89.48\% \\
        ColaGNN   & 2566 & 5746 &50.29\%  & 92.17\% &82.16\%  & 5294 &  10402& 101.50\% &86.60\%  & 63.78\% \\
        STAN &1070  &  2400&22.97\%  & 95.87\% &94.82\%  &\underline{1623}  & 3165 & \underline{34.38\%} & 94.80\%& 91.97\%  \\
        MPSTAN  &  \underline{1016}& 2311 & \textbf{16.91\%} &96.74\%  &95.60\%  &\textbf{1356}  &  \underline{3016}& \textbf{24.34\%}&93.38\%  & \underline{92.27\%} \\
        Improvement &-  & - & 6.37\% & - & - & 16.45\% & - &29.20\%  & - &-  \\
        \toprule
        \multirow{4}*{Model}&\multicolumn{5}{l}{T=15}&\multicolumn{5}{l}{T=20}\\
        \cmidrule(l){2-6}\cmidrule(l){7-11}
          & MAE & RMSE &MAPE & PCC & CCC  & MAE & RMSE &MAPE & PCC & CCC  \\
        \midrule 
        SIR  & 2632 & 4373 &66.60\%  & \textbf{95.22\%} & 87.05\% & 3515 & 5883 &92.93\%  &\underline{92.08\% } &   79.20\% \\
        ARIMA & 3443 & 7715 & 86.16\% & 65.62\% & 61.39\% &3757  & 7513 &130.90\%  &72.79\%  & 66.56\% \\
        GRU   &  2124&\underline{3758}  &59.84\%  & 88.58\% &  87.70\%& 2977 &5343  &68.13\%  &  71.75\%&  70.72\%  \\
        GraphWaveNet &2828  &  6520&  \underline{49.39\%}& 93.62\% & 79.34\% & 2773 & 6547  & \underline{46.11\%} &  \textbf{92.96\%}& 79.38\% \\
        STGODE &10330  &23345  &195.76\% & 82.22\% & 38.62\% & 12156 & 27407 & 221.51\% & 83.58\% & 33.33\%   \\
        CovidGNN  &2988  &  6515& 66.73\% & 90.20\% & 77.42\% &3990  &  8805& 94.82\% &84.97\%  & 67.12\% \\
        ColaGNN    &4192  & 8688 & 93.21\% &84.31\%  & 67.68\% & 7195 &15400  & 140.32\% & 84.30\% & 50.40\%  \\
        STAN & \underline{2026} & 3887 & 51.03\% & \underline{93.86\%} & \underline{88.92\%} & \underline{2804} & \underline{5238} & 72.10\% & 90.59\%& \underline{82.24\%}  \\
        MPSTAN &  \textbf{1465}& \textbf{3104 }&\textbf{28.29\% } & 91.84\% & \textbf{91.29\%} & \textbf{1854} & \textbf{4014} &\textbf{34.67\% } &  85.78\%& \textbf{84.97\% } \\
        Improvement &27.69\%  & 17.40\% &  42.72\%& - & 2.67\% & 33.88\%&  23.37\%& 24.81\% &-  &  3.32\%\\
          \bottomrule
    \end{tabular}
    }
    \caption{Performance comparison with baseline on the Japan dataset.}
    \label{table:japan-exp}
\end{table*}
\subsection{Forecasting Performance}
As shown in \reftable{table:us-exp} and \reftable{table:japan-exp}, we evaluate the performance of our method with all the baselines on the the US dataset and the Japan dataset for predicting daily active cases, respectively, where bolded and underlined indicate optimal and suboptimal, and Improvement denotes the improved rate of MPSTAN compared to the suboptimal forecasting results. On the US dataset, our method achieves state-of-the-art (SOTA) performance for both short-term (T=5,10) and long-term (T=15,20) forecasting. In particular, our forecasting results for all the forecasting tasks show significant improvements over the suboptimal forecasting, where MAE improves at least 19.05\%, RMSE improves at least 7.72\%, MAPE improves at least 23.97\%, PCC improves at least 0.34\%, and CCC improves at least 0.11\%. While our method may not fully achieve the SOTA performance on the Japan dataset, it can achieve optimal or competitive forecasting results compared to other models, demonstrating strong competitiveness, where MAE improves at least 16.45\%, RMSE improves at least 17.40\%, MAPE improves at least 6.38\%, and CCC improves at least 2.66\%. In summary, compared to all baseline models, MPSTAN can provide more accurate and stable forecasting for different real-world epidemic datasets.
\begin{table*}
    \centering
    \resizebox{\textwidth}{!}{
    \begin{tabular}{lllllllllll}
      \toprule
        &\multicolumn{10}{l}{The US dataset}\\
        \cmidrule(l){2-11}
        &\multicolumn{5}{l}{T=5}&\multicolumn{5}{l}{T=10}\\
        \cmidrule(l){2-6}\cmidrule(l){7-11}
         Model & MAE & RMSE &MAPE & PCC & CCC  & MAE & RMSE &MAPE & PCC & CCC  \\
        \midrule 
        MPSTAN w/o Phy-All  &14865  & 34756 &10.96\%  & 96.19\% & 95.18\% &  22911&54185  & 17.38\% & 91.55\% &  86.63\%  \\
        MPSTAN w/o Phy-Loss & 18908 & 39201&14.62\%  &  94.53\%& 92.97\% & 15201 & 27700 & 16.09\% &98.53\%  &  96.57\%  \\
        MPSTAN w/o Phy-Model &19002  & 45127 & 13.04\% &94.09\% & 90.52\% & 25372 & 64364 &18.17\%  & 86.59\% &  81.28\% \\
        MPSTAN w/o Mobility &  5030&9845  & 7.09\% & 99.78\% & 99.65\% &8147  & 14895 &11.17\%  &\textbf{99.56\%}  &  99.16\% \\
        MPSTAN w/o MPG&4399  &9033  &6.71\%  & 99.77\% & 99.70\% &\textbf{7640}  & \textbf{14456} & \textbf{10.70\%} & 99.55\% &    \textbf{99.21\%}\\
        MPSTAN & \textbf{3960} & \textbf{8255} & \textbf{6.38\%} & \textbf{99.80\%} & \textbf{99.75\% }&7711  & 14463 &  10.73\%& 99.55\% & 99.20\% \\
        \toprule
        &\multicolumn{5}{l}{T=15}&\multicolumn{5}{l}{T=20}\\
        \cmidrule(l){2-6}\cmidrule(l){7-11}
          Model& MAE & RMSE &MAPE & PCC & CCC  & MAE & RMSE &MAPE & PCC & CCC  \\
        \midrule 
        MPSTAN w/o Phy-All  &22876  & 58160 & 19.10\% &88.68\%  & 85.34\% & 27659 & 60632 &  24.83\%& 89.30\% & 83.16\%  \\
        MPSTAN w/o Phy-Loss & 18526 & 32033 & 20.19\% & 99.15\% &  95.58\%&  22138&37753  & 24.06\% & 97.54\% & 93.59\%  \\
        MPSTAN w/o Phy-Model &  27509& 63056 &  21.84\%& 88.68\% &80.58\%  &  27425& 61194 &24.36\%  & 88.96\% & 82.69\%  \\
        MPSTAN w/o Mobility & 11054 & 20240& 15.33\% & 99.19\% &98.39\%  & \textbf{11859} &\textbf{22477}  & \textbf{18.37\%} & 98.71\% &   \textbf{98.02\%} \\
        MPSTAN w/o MPG & 10441 & 18984 & 14.92\% & \textbf{99.25\%} &98.59\%  &  13064& 23702 & 18.98\% &\textbf{98.87\%}  &  97.75\%  \\
        MPSTAN &  \textbf{10148}&  \textbf{18460}& \textbf{14.68\%} & \textbf{99.25\%} &  \textbf{98.68\% } & 12728 & 22923 & 18.68\% & 98.81\% & 97.91\%  \\
          \bottomrule
    \end{tabular}
    }
    \caption{Ablation study on the US dataset.}
    \label{table;ablation-us}
\end{table*}
\begin{table*}
    \centering
    \resizebox{\textwidth}{!}{
    \begin{tabular}{lllllllllll}
      \toprule
        &\multicolumn{10}{l}{The Japan Dataset}\\
        \cmidrule(l){2-11}
        &\multicolumn{5}{l}{T=5}&\multicolumn{5}{l}{T=10}\\
        \cmidrule(l){2-6}\cmidrule(l){7-11}
         Model & MAE & RMSE &MAPE & PCC & CCC  & MAE & RMSE &MAPE & PCC & CCC  \\
        \midrule 
        MPSTAN w/o Phy-All  &3326  & 10410 &26.59\%  &  92.27\%& 66.80\% & 3201 & 9632 & 29.52\% & 92.06\% &   68.65\%  \\
        MPSTAN w/o Phy-Loss & \textbf{928} & \textbf{2024} & \textbf{15.81\%} &96.05\%  & \textbf{95.90\%} & \textbf{1196} & \textbf{2620} & \textbf{22.31\%} & \textbf{93.70\%} &\textbf{93.50\% }   \\
        MPSTAN w/o Phy-Model &3674  & 11714 &  28.11\%& 91.84\% & 62.93\% & 3896 & 10762 & 45.96\% & 91.48\% & 64.93\%  \\
        MPSTAN w/o Mobility &1142  &\textbf{2309}  &  21.93\%& \textbf{98.46\%}& 95.60\% & \textbf{1273} & \textbf{2633} &27.24\%  & \textbf{96.77\%} & \textbf{94.33\%}  \\
        MPSTAN w/o MPG& 1047 & 2339 & 19.18\% &96.52\%  & 95.44\% & \textbf{1216} &  \textbf{2630}&\textbf{23.49\% }& \textbf{94.52\% }&   \textbf{93.90\%} \\
        MPSTAN&  1016& 2311 & 16.91\% &96.74\%  &95.60\%  &1356  &  3016& 24.34\% &93.38\%  & 92.27\% \\
        \toprule
        &\multicolumn{5}{l}{T=15}&\multicolumn{5}{l}{T=20}\\
        \cmidrule(l){2-6}\cmidrule(l){7-11}
          Model& MAE & RMSE &MAPE & PCC & CCC  & MAE & RMSE &MAPE & PCC & CCC  \\
        \midrule 
        MPSTAN w/o Phy-All  &3435  & 9796 & 35.25\% & 91.37\% &67.63\%  & 3054 & 7736 &  41.09\%& \textbf{90.29\%} &  74.42\%  \\
        MPSTAN w/o Phy-Loss & 1774 & 3941 & 32.28\% &84.22\%  &  83.44\%& 1928 &  4383& 41.30\% & \textbf{86.70\%} &  84.83\% \\
        MPSTAN w/o Phy-Model &3897  & 11099 & 42.38\% &  90.35\%& 63.40\% & 4273 &  10664& 69.34\% &\textbf{86.83\%}  & 63.25\%   \\
        MPSTAN w/o Mobility & \textbf{1100} & \textbf{2271} & \textbf{24.42\%} & \textbf{96.25\%} &\textbf{95.31\%} & \textbf{1319} &\textbf{2958}  & \textbf{26.72\%} &  \textbf{93.80\%}& \textbf{92.42\%}   \\
        MPSTAN w/o MPG & \textbf{1391} & 3379 & \textbf{25.48\%} & \textbf{91.89\%} &90.44\%  & \textbf{1786} &4073  &  \textbf{31.19\%}& \textbf{87.89\%}& \textbf{85.75\%}\\
        MPSTAN &  1465& 3104 &28.29\%  & 91.84\% & 91.29\% & 1854 & 4014 &34.67\%  &  85.78\%& 84.97\%  \\
          \bottomrule
    \end{tabular}
    }
    \caption{Ablation study on the Japan dataset.}
    \label{table;ablation-japan}
\end{table*}

\par Next, we discuss specifically the performance comparison between different models. Traditional mathematical models (e.g., SIR, ARIMA) often outperform neural network models in short-term forecasting, but the performance becomes worse in long-term forecasting. This may be because the predictive accuracy of traditional mathematical models is highly dependent on the time length, and long-term forecasting requires more historical data. Insufficient historical data can lead to forecasting errors, and the cumulative effect of errors increases with longer forecasting times, resulting in worse long-term forecasting results. 
\par In addition, we observe that traffic flow models, particularly the STGODE, face challenges in providing stable and accurate forecasting for different tasks. This may be attributed to the fact that epidemic data is sparser and noisier than traffic flow data, increasing the likelihood of these models overfitting when applied to epidemic data. Through observation, it is noticed that the ColaGNN model also faces difficulties in providing accurate forecasting. It is believed that the ColaGNN model was originally designed for influenza-like illnesses, while COVID-19 data is more complex and on a larger scale. As a result, the ColaGNN model is not well-suited for these tasks.
\par By comparing domain knowledge-based models (e.g., STAN, MPSTAN) with other baselines, we observe that STAN and MPSTAN outperform other models in terms of accuracy, indicating that neural networks incorporating epidemiological domain knowledge better capture the underlying dynamics of epidemic transmission and achieve more accurate forecasting. In particular, the results show that MPSTAN performs better than STAN, highlighting the value of this integrated neural network framework that combines epidemiological domain knowledge to achieve more accurate forecasting. This framework involves two main aspects: integrating domain knowledge and modeling metapopulation transmission. Furthermore, in section \ref{sec:Ablation Study}, we will discuss the impact of these two aspects on forecasting results, including the effects of integration methods and inter-patch interactions.
\begin{figure*}[t]
	\centering  
	\subfigure[US active cases]{
		\includegraphics[width=0.48\linewidth]{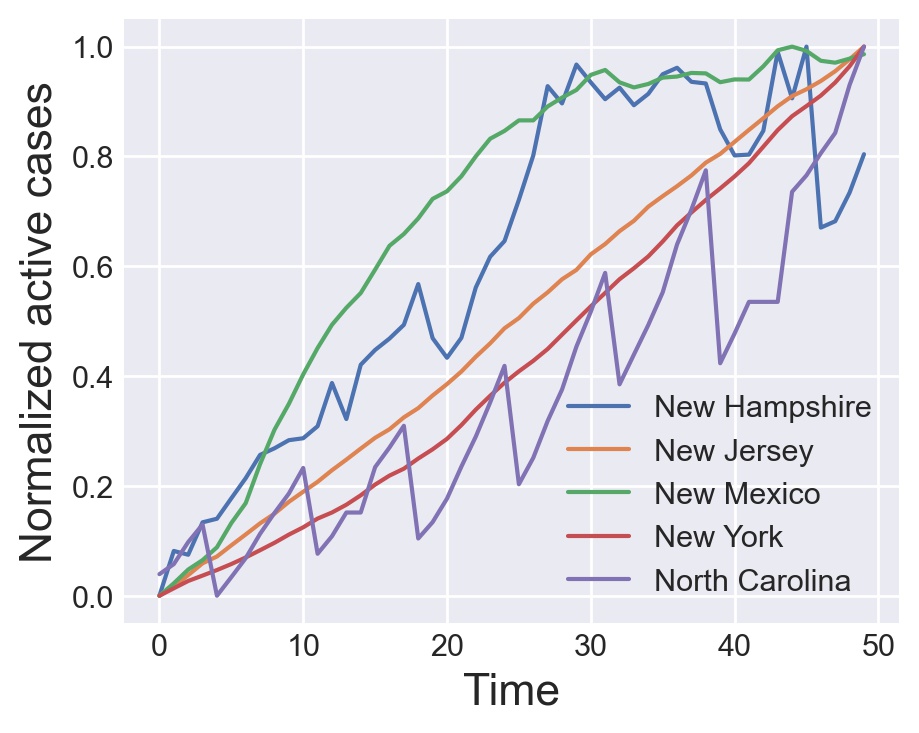}}
	\subfigure[Japan active cases]{
		\includegraphics[width=0.48\linewidth]{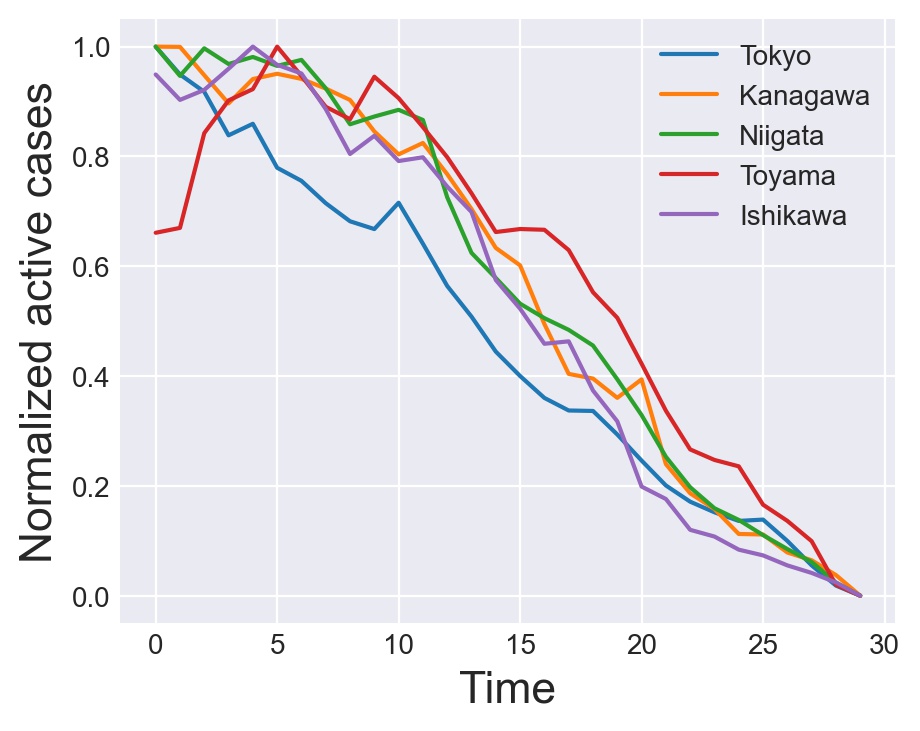}}
	\caption{Samples of typical cities in the US and Japan datasets.}
    \label{fig:test}
\end{figure*}
\begin{figure*}[t]
	\centering  
	\subfigure[GRU-MAE]{
		\includegraphics[width=0.32\linewidth]{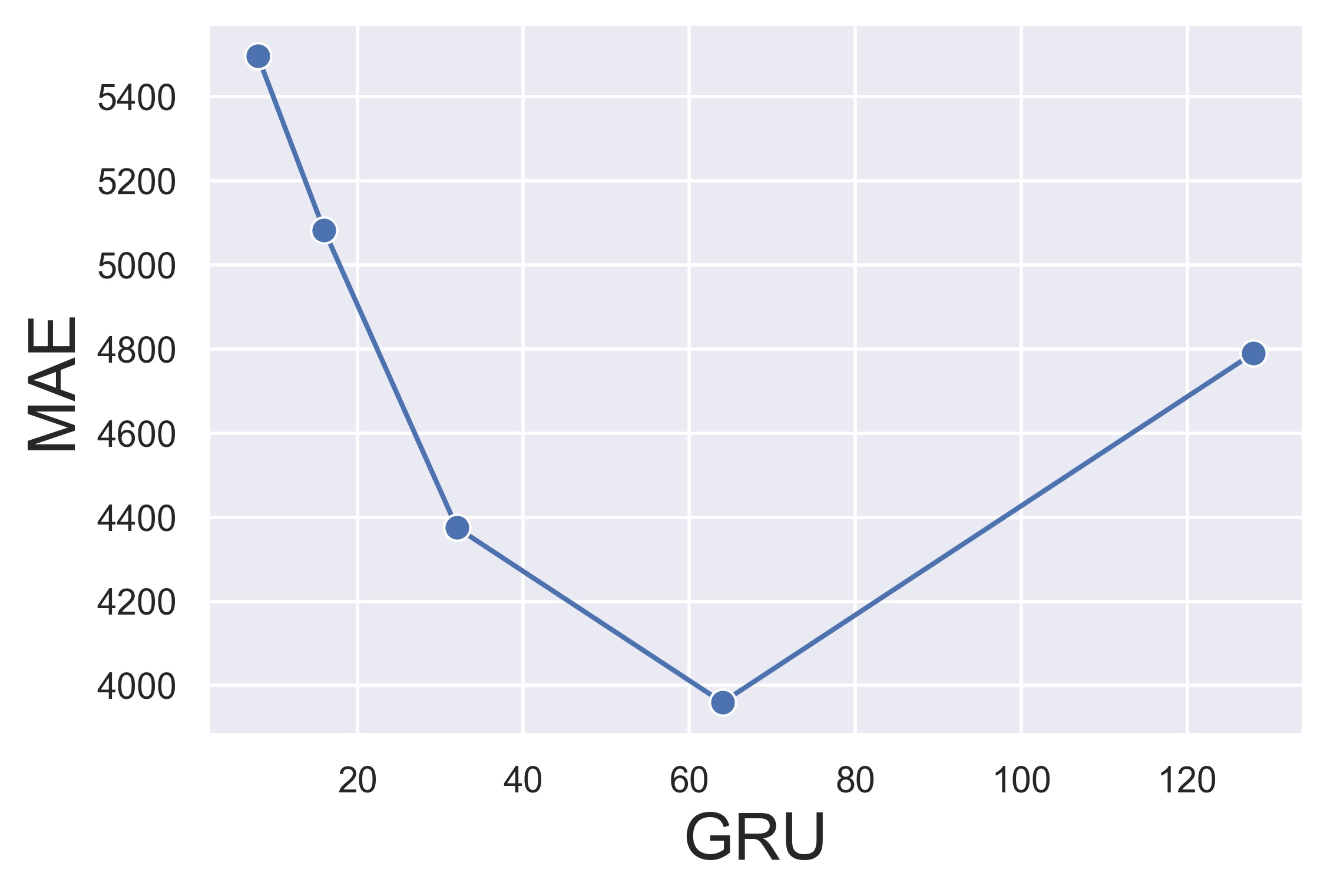}}
        \subfigure[GRU-RMSE]{
		\includegraphics[width=0.32\linewidth]{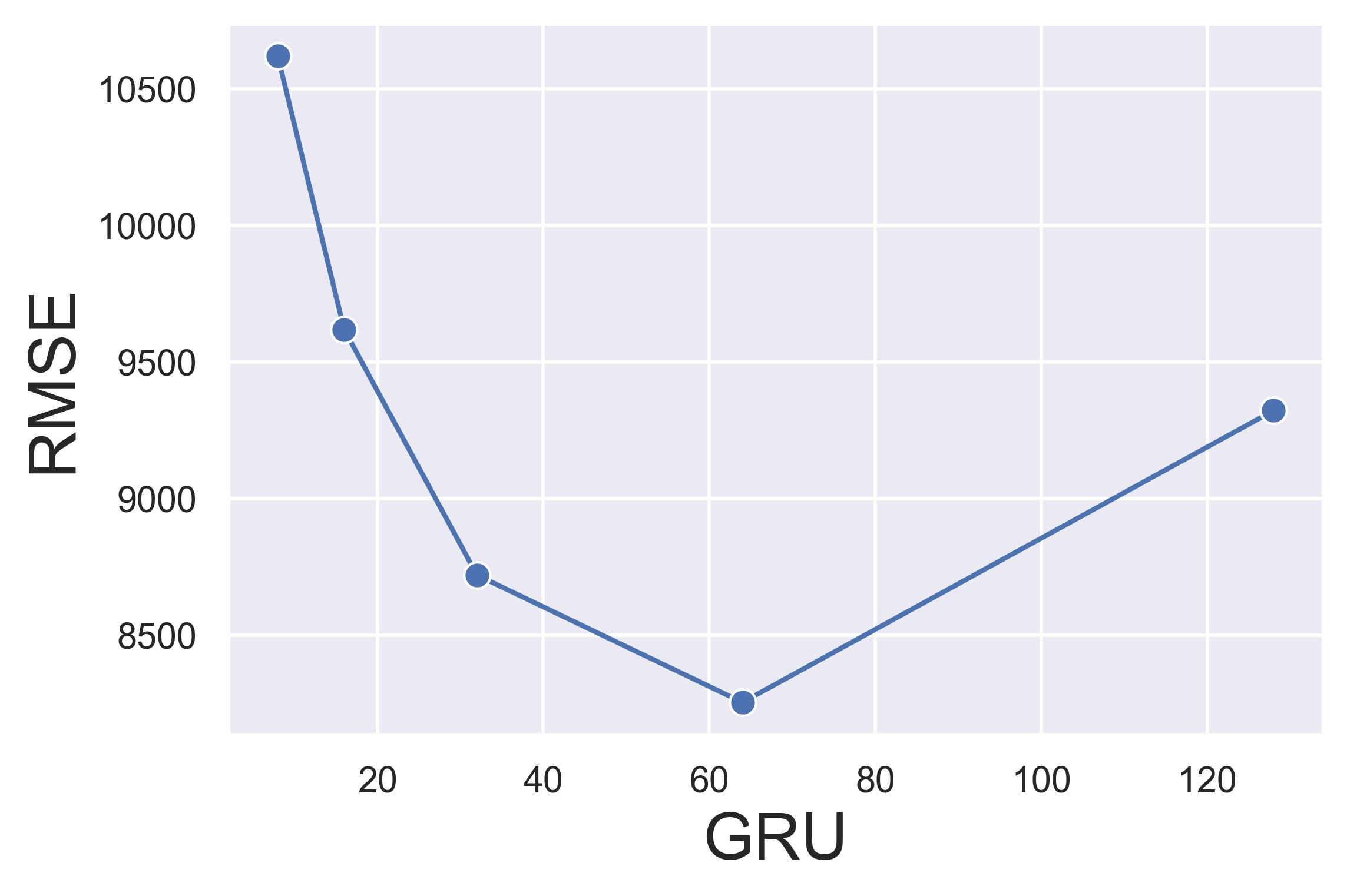}}
	\subfigure[GRU-MAPE]{
		\includegraphics[width=0.32\linewidth]{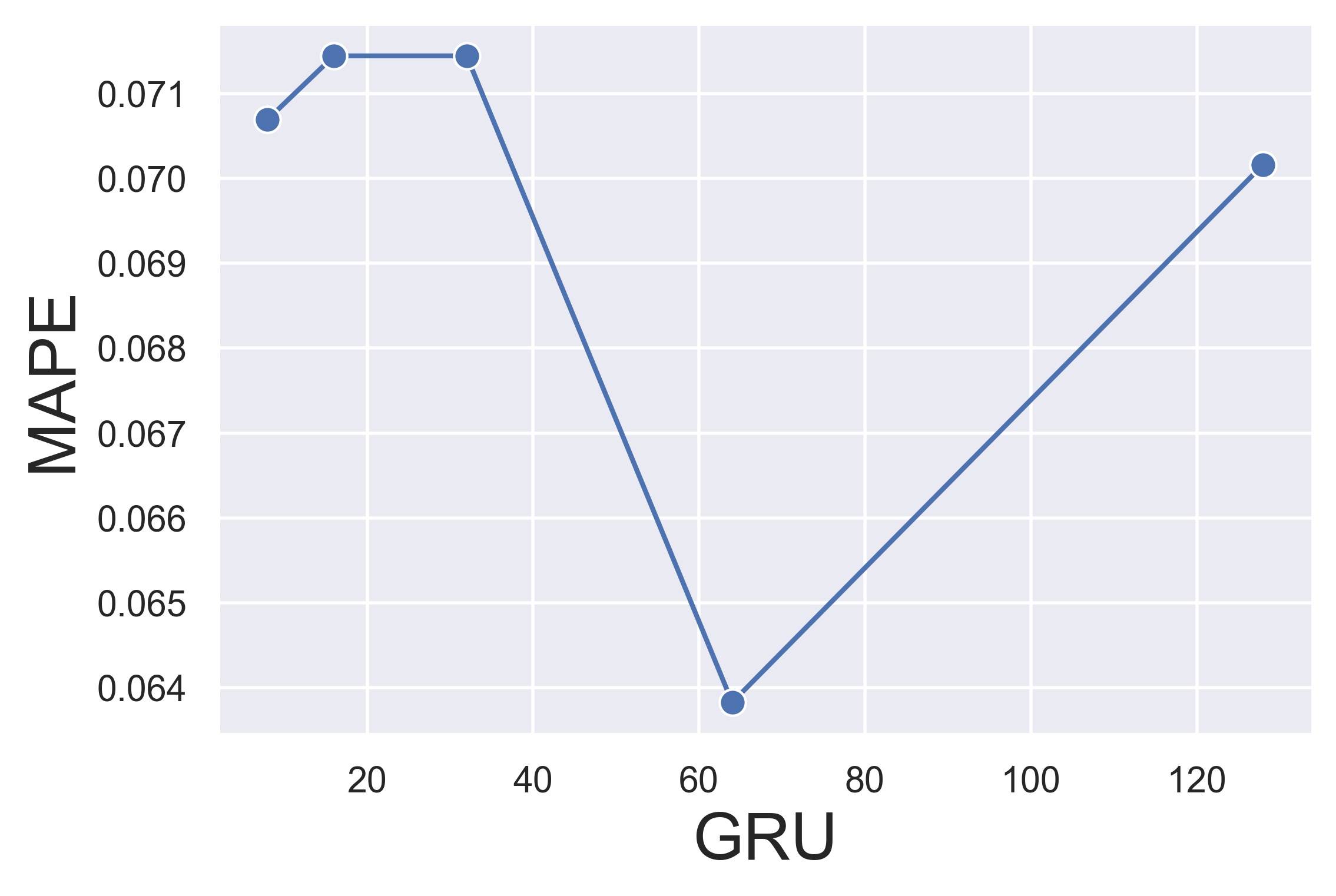}}
        \subfigure[GAT-MAE]{
		\includegraphics[width=0.32\linewidth]{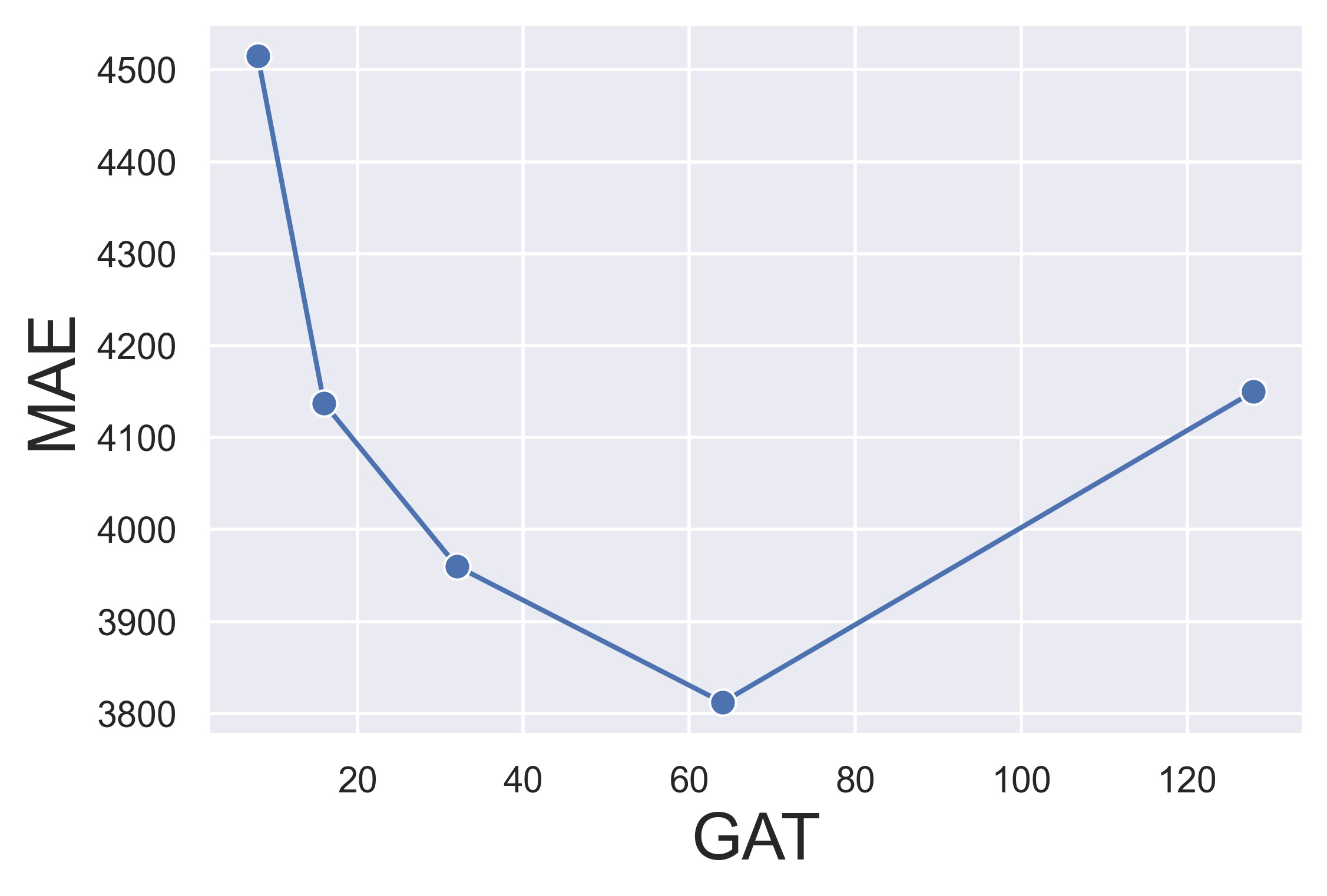}}
        \subfigure[GAT-RMSE]{
		\includegraphics[width=0.32\linewidth]{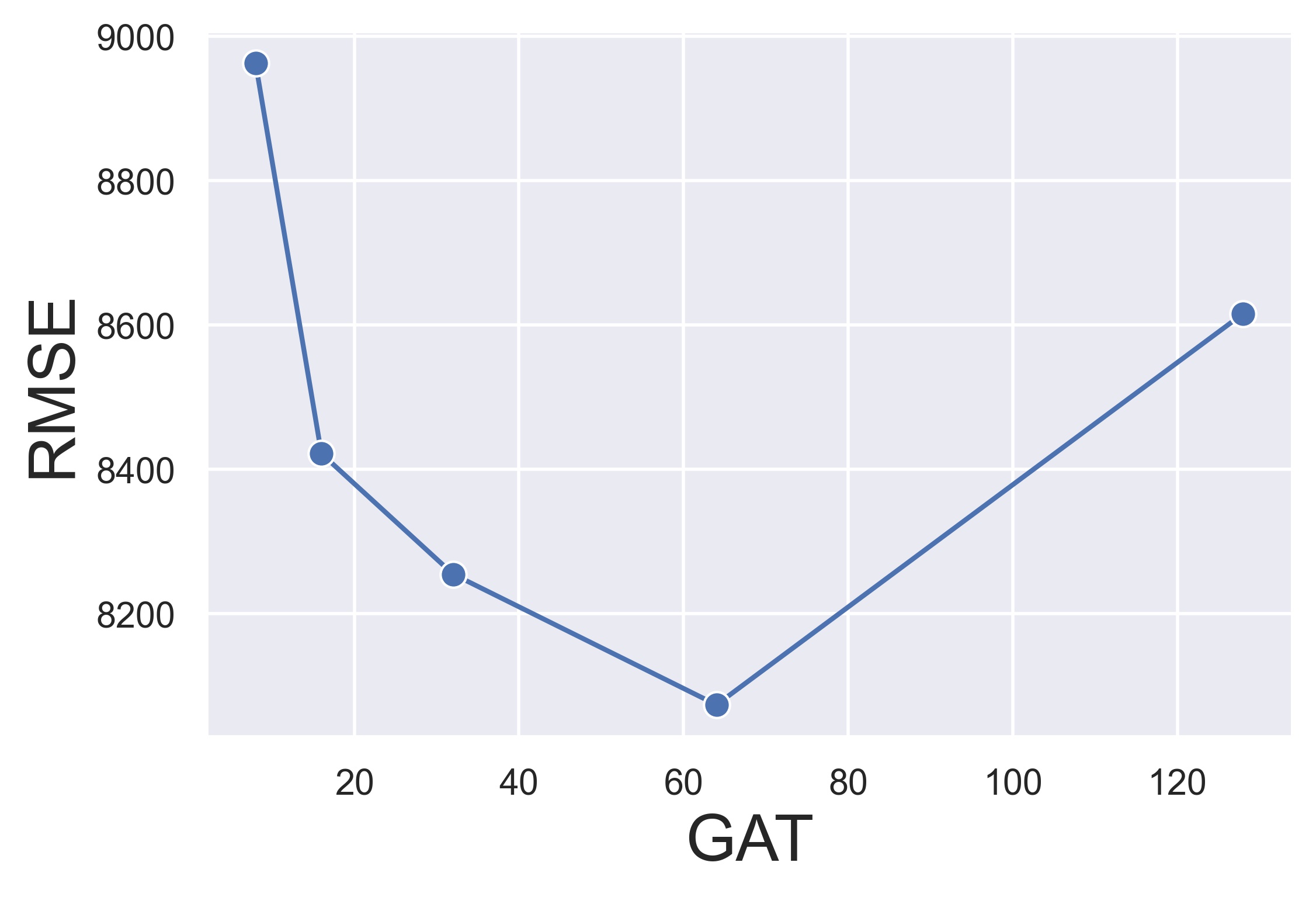}}
	\subfigure[GAT-MAPE]{
		\includegraphics[width=0.32\linewidth]{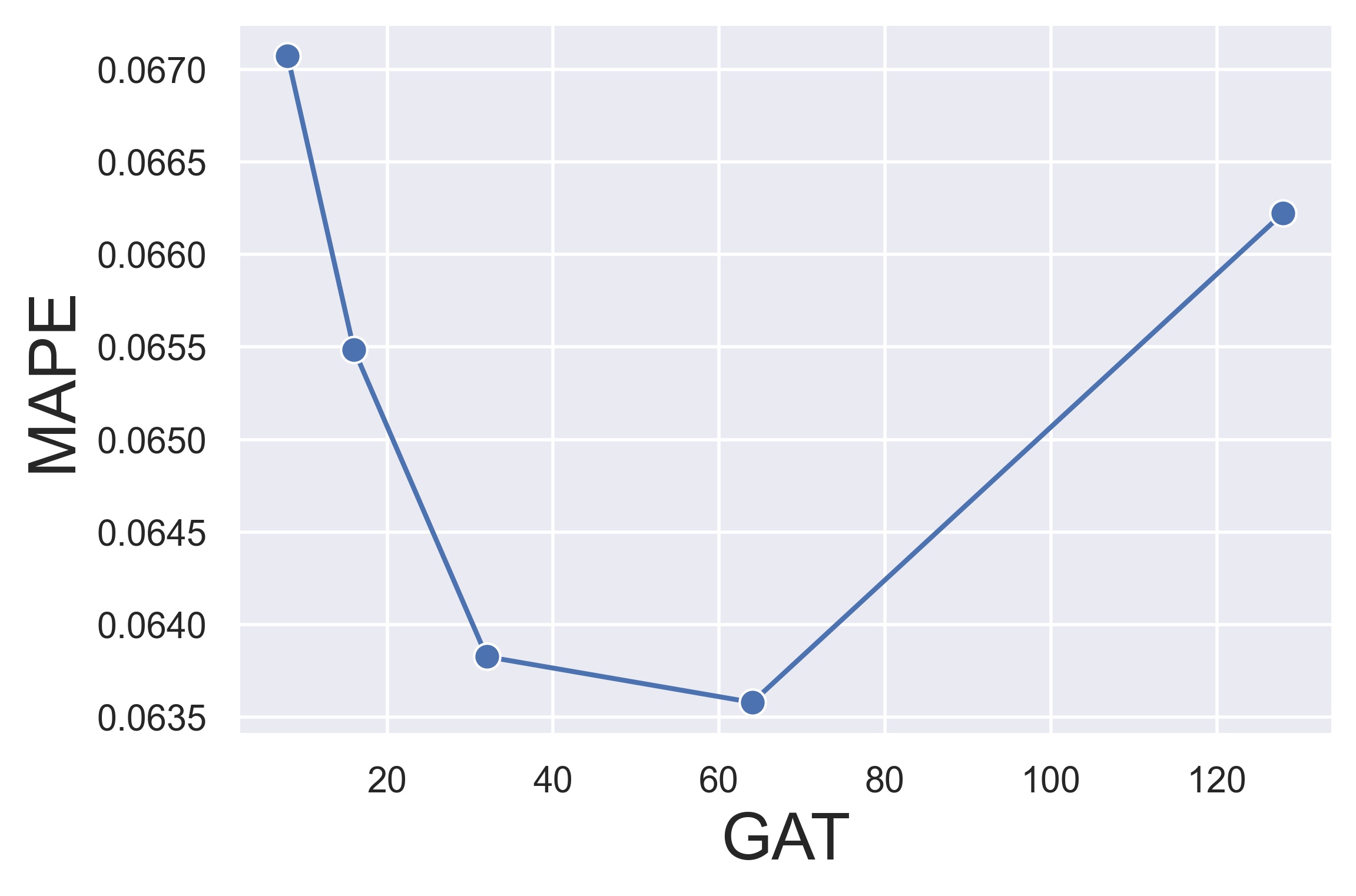}}
	\caption{Effect of hyperparameters on performance.}
    \label{fig:test-para-1}
\end{figure*}
\begin{figure}[t]
	\centering   
	\includegraphics[width=\linewidth]{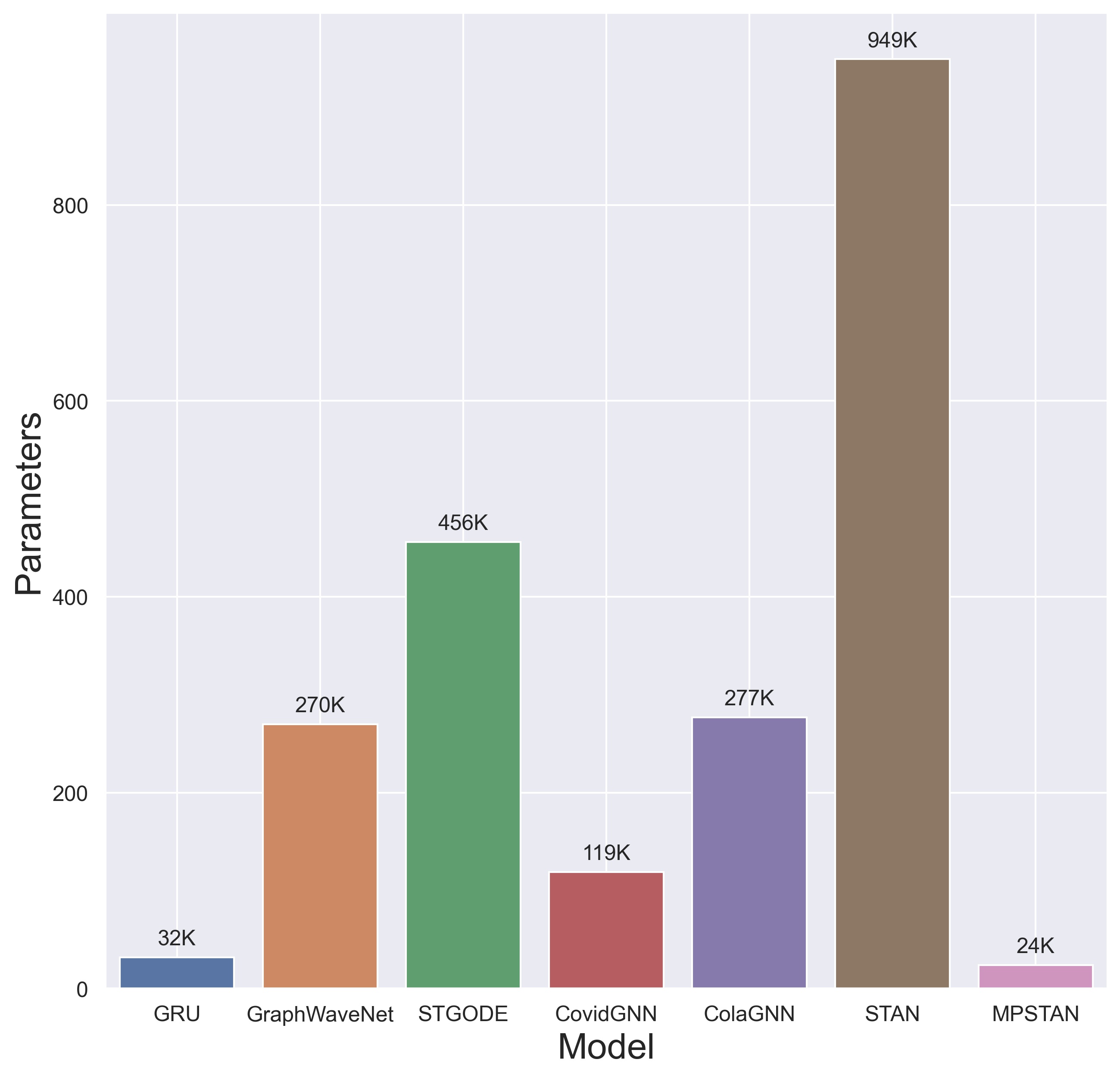}
	\caption{Comparison of model complexity.}
        \label{fig:para}
\end{figure} 
\subsection{Ablation Study}\label{sec:Ablation Study}
To explore the impact of epidemiological domain knowledge on epidemic forecasting and to verify the effectiveness of the model components, we further conduct ablation experiments on the US and Japan datasets.
\begin{enumerate}[(1)]
  \item \textbf{MPSTAN w/o Phy-All}: Remove epidemiological domain knowledge from both the model construction and loss function. We use only the spatio-temporal module for epidemic forecasting.
  \item \textbf{MPSTAN w/o Phy-Loss}: Remove epidemiological domain knowledge from the loss function. We only combine the  knowledge into the model construction.
  \item \textbf{MPSTAN w/o Phy-Model}: Remove the epidemiological domain knowledge from the model construction. We predict physical model parameters in the output layer and combine the knowledge into the loss function.
  \item \textbf{MPSTAN w/o Mobility}: Combine epidemiological domain knowledge without considersing population mobility into the model, mainly by using the SIR model instead of the MP-SIR model.
  \item \textbf{MPSTAN w/o MPG}: Remove multiple parameter generators (MPG). We generate all the physical model parameters using a single parameter generator for embeddings containing spatio-temporal information.
\end{enumerate}

\par The results of the ablation experiments are shown in \reftable{table;ablation-us} and \reftable{table;ablation-japan}, where bold indicates better performance for the ablation model or MPSTAN. Firstly, we analyze the effectiveness of domain knowledge in epidemic forecasting by comparing the performance of the MPSTAN with MPSTAN w/o Phy-All on two datasets. The results show that the MPSTAN w/o Phy-All model, which lacks domain knowledge, performs extremely poorly in epidemic forecasting, highlighting the crucial role of epidemiological domain knowledge in epidemic forecasting. 
\par To further investigate the impact of different methods for integrating domain knowledge on epidemic forecasting, we compare MPSTAN w/o Phy-Loss, MPSTAN w/o Phy-Model with MPSTAN. On the US dataset, MPSTAN, which applies domain knowledge to both model construction and loss function, can more accurately predict epidemic trends, as shown in \reftable{table;ablation-us}. In \reftable{table;ablation-japan}, for short-term forecasting on the Japan dataset, MPSTAN performs worse than MPSTAN w/o Phy-Loss, which only applies domain knowledge to model construction, but still provides competitive forecasting. In long-term forecasting, MPSTAN outperforms the other two models. Overall, incorporating domain knowledge into both model construction and loss function can better help the model learn the basic dynamics of epidemic transmission and improve forecasting accuracy. By comparing MPSTAN w/o Phy-Loss and MPSTAN w/o Phy-Model on two datasets, we find that the former performs better in all forecasting tasks, indicating that applying domain knowledge to model construction is more beneficial for accurate epidemic forecasting than applying it to the loss function. In addition, by comparing MPSTAN w/o Phy-All and MPSTAN w/o Phy-Model, we find that using domain knowledge to only constrain the loss function may lead to poorer forecasting performance. Therefore, we believe that incorporating domain knowledge into model construction is essential, and simultaneously applying it to the loss function can improve the predictive accuracy of the model.
\par For the remaining model components, the effectiveness of the metapopulation model establishment and multiple parameter generators can be verified by using MPSTAN w/o Mobility and MPSTAN w/o MPG, respectively. On the US dataset, MPSTAN outperforms MPSTAN w/o Mobility for forecasting tasks with T=5, 10, and 15. However, the opposite result is observed for the T=20 task, which may be due to the fact that inter-patch physical parameters are no longer sufficient to define the population mobility when the forecasting time is longer. Overall, MP-SIR, a metapopulation epidemic model that considers population mobility, is more beneficial for model training than traditional SIR. Additionally, comparing MPSTAN with MPSTAN w/o MPG reveals that using only one parameter generator to generate all physical model parameters may lead to poorer predictive performance. 
\par On the Japan dataset, we observe that the performance of MPSTAN w/o Mobility and MPSTAN w/o MPG is mostly superior to MPSTAN. We believe that this is due to the fact that these two datasets are collected at different times and locations, leading to differences in disease control measures and public awareness. To confirm this, we randomly select five cities from each dataset and display the normalized daily active cases of these cities in \reffig{fig:test}. It clearly shows that US cities are experiencing a surge in active cases, while Japan cities are effectively controlling the spread of the disease, resulting in a decrease in active cases. Moreover, we investigate the Covid-19 Community Mobility Reports~\cite{aktay2020google} from Google for the corresponding time periods of these two datasets. We observe that the park population movement in the US is higher than the pre-epidemic baseline, while in Japan it is lower than the baseline. Possible reasons for the above situation could be that the data collected in the United States is from an earlier period when the COVID-19 prevention and control policies are not yet well-established, resulting in greater population mobility. On the other hand, the data collected in Japan is from a later period when more comprehensive measures have been implemented and the public has become more aware of the importance of self-isolation, leading to lower population mobility. Therefore, on the Japan dataset, the traditional SIR model is more suitable to be combined with neural networks for epidemic forecasting. The multiple parameter generators (MPG) are essentially based on the metapopulation epidemic model, and thus, the forecasting accuracy of MPSTAN w/o MPG is higher. 
\par Furthermore, we recognize that no single domain knowledge can be universally applied to all complex epidemic data. Thus, when selecting domain knowledge to integrate into neural networks, it is necessary to consider the actual circumstances and choose more representative knowledge to achieve more accurate forecasting.

\subsection{Effect of Hyperparameters}
\par In this section, we study the effect of hyperparameters on performance, focusing on the dimensions of GRU and GAT. We vary one parameter at a time while keeping the other parameter constant. In addition, the dimension range is set to [8, 16, 32, 64, 128], T=5 is selected as the task on the US dataset, while MAE, RMSE, and MAPE are chosen as the evaluation metrics.
\par \reffig{fig:test-para-1} shows the effects of different dimensions of GRU and GAT on the performance, respectively. It can be seen that the forecasting performance is poor when the number of dimensions is small, and gradually becomes better when the number of dimensions increases, which is because more parameters are involved in fitting the potential dynamics of the epidemic. When the number of dimensions continues to increase, the forecasting performance will also become worse. The possible reason of this issue may be that the epidemic data are sparse and the excessive number of parameters will lead to the overfitting problem.

\subsection{Model Complexity}
\par We analyze the model complexity by comparing the neural network parameters of all models. As shown in \reffig{fig:para}, the number of neural network parameters in MPSTAN is significantly less than in other spatio-temporal models. This is because MPSTAN makes extensive use of epidemiological domain knowledge (e.g., model construction, loss functions), thus reducing the reliance on neural networks and lowering the number of parameters. By comparing GRU and MSPTAN, we find that the number of parameters is similar, but the former ignores the spatial dependence and the intrinsic propagation mechanism of the epidemic which can only be used for temporal forecasting of a single patch, while the latter perfectly solves the above problems and provides stable and accurate forecasting for different trends.
\section{Conclusion} \label{sec:Conclusion}
In this paper, we propose a Metapopulation-based Spatio-Temporal Attention Network (MPSTAN) for epidemic forecasting. The model uses an adaptive approach to define interactions between patches and applies the constructed domain model to model construction and loss function of MPSTAN to better learn the underlying dynamics of epidemic propagation. Experiments show that the MPSTAN outperforms other baselines and is more stable on two real datasets with different epidemiological evolution trends. Additionally, we further analyze the effectiveness of incorporating domain knowledge and find that it improves the accuracy of forecasting in the learning model. Specifically, domain knowledge plays a more critical role in model construction than loss functions, and applying it to both aspects can better fit to potential epidemiological dynamics. We also recognize that no single domain knowledge can perfectly fit epidemic forecasting in different real-world situations. Instead, we should select domain knowledge that is more representative based on the actual circumstances to achieve more accurate forecasting. We also discuss the impact of hyperparameters on the model, as excessively small or large hyperparameters can lead to underfitting or overfitting, respectively, so appropriate hyperparameters must be chosen. Finally, we analyze the model complexity and find that compared to all baselines, MPSTAN requires fewer neural network parameters due to its greater integration of domain knowledge.
\par Our model achieves state-of-the-art or competitive results in epidemic forecasting for different epidemic trends, but there are still several aspects where performance can be improved. Firstly, graph construction has a significant impact on the entire learning model, as it affects the propagation of spatial information and the inter-patch interactions of the physical model. Therefore, a reasonable graph structure is crucial. Currently, we use the gravity model to construct the graph structure, which relies on prior knowledge, but may overlook some potential information, resulting in an incomplete capture of the correct graph information between patches. In addition, the graph information between patches changes over time, rather than being fixed. Hence, in the future, we will combine potential graph information to construct a dynamic graph structure to better describe the interactive graph of epidemics. Furthermore, in the model construction, we currently simply connect the neural network results with domain knowledge from physical model without considering their respective roles or weights, which may also lead to a decrease in accuracy. Therefore, we will carefully analyze the roles of the neural network and domain knowledge in epidemic forecasting and explore more effective methods to fuse the information of the two, such as introducing gating mechanisms to achieve more accurate forecasting.




\printcredits
\vskip 0.7 cm\noindent\textbf{Declaration of competing interest}

The authors declare that they have no known competing financial interests or personal relationships that could have appeared to influence the work reported in this paper.

\vskip 0.7 cm\noindent\textbf{Acknowledgments}

This work was supported by the National Natural Science Foundation of China (Grant No. 52273228), Natural Science Foundation
of Shanghai, China(Grant No. 20ZR1419000), Key Research Project of
Zhejiang Laboratory (Grant No. 2021PE0AC02).
\bibliographystyle{model1-num-names}
\bibliography{references}
\end{document}